\documentclass[acmsmall]{acmart}

\usepackage{booktabs} 
\usepackage{hyperref}  
\usepackage{url} 
\usepackage{multirow}  
\usepackage{color}

\usepackage{appendix}
\usepackage{lineno}
\usepackage{diagbox}
\usepackage{amsmath}
\usepackage{arydshln} 
\usepackage{algorithmic}

\usepackage{tabularx}
\usepackage{makecell}

\usepackage[linesnumbered,ruled]{algorithm2e} 

\SetAlFnt{\small}
\SetAlCapFnt{\small}
\SetAlCapNameFnt{\small}
\SetAlCapHSkip{0pt}
\IncMargin{-\parindent}
\usepackage{longtable}

\usepackage{soul}
\usepackage{color, xcolor} 
\sethlcolor{yellow}
\soulregister{\cite}7 
\soulregister{\citep}7 
\soulregister{\citet}7 
\soulregister{\ref}7 
\soulregister{\pageref}7 
\soulregister{\underline}7 

\acmJournal{TIST}

\setcopyright{acmcopyright}

\begin{document}


\title{Graph Attention-based Adaptive Transfer Learning for Link Prediction}

\author{Huashen Lu}
\affiliation{
	\institution{Jinan University}
	\city{Guangzhou}
	\country{China}
}
\email{huashen.lu@gmail.com}

\author{Wensheng Gan}
\authornote{This is the corresponding author.}
\affiliation{
	\institution{Jinan University}
	\city{Guangzhou}
	\country{China}
}
\email{wsgan001@gmail.com}

\author{Guoting Chen}
\affiliation{ 
	\institution{Great Bay University}
	\city{Dongguan}
	\country{China}
}
\email{guoting.chen@univ-lille.fr}

\author{Zhichao Huang}
\affiliation{ 
	\institution{JD Technology}
	\city{Beijing}
	\country{China}
}
\email{iceshzc@gmail.com}

\author{Philip S. Yu}
\affiliation{
	\institution{University of Illinois Chicago}
	\city{Chicago}
	\country{USA}
}
\email{psyu@uic.edu}

\begin{abstract}
  Graph neural networks (GNNs) have brought revolutionary advancements to the field of link prediction (LP), providing powerful tools for mining potential relationships in graphs. However, existing methods face challenges when dealing with large-scale sparse graphs and the need for a high degree of alignment between different datasets in transfer learning. Besides, although self-supervised methods have achieved remarkable success in many graph tasks, prior research has overlooked the potential of transfer learning to generalize across different graph datasets. To address these limitations, we propose a novel Graph Attention Adaptive Transfer Network (GAATNet). It combines the advantages of pre-training and fine-tuning to capture global node embedding information across datasets of different scales, ensuring efficient knowledge transfer and improved LP performance. To enhance the model's generalization ability and accelerate training, we design two key strategies: 1) Incorporate distant neighbor embeddings as biases in the self-attention module to capture global features. 2) Introduce a lightweight self-adapter module during fine-tuning to improve training efficiency. Comprehensive experiments on seven public datasets demonstrate that GAATNet achieves state-of-the-art performance in LP tasks. This study provides a general and scalable solution for LP tasks to effectively integrate GNNs with transfer learning. The source code and datasets are publicly available at \url{https://github.com/DSI-Lab1/GAATNet}.
\end{abstract}

%
%



\maketitle

\renewcommand{\shortauthors}{H. Lu \textit{et al.}}

\section{Introduction}  \label{sec: introduction}

In recent years, graph-structured data has found wide applications in various domains, including social networks, biological systems, and recommendation engines. Link prediction (LP), a fundamental task in graph data mining, aims to predict potential edges based on existing network topology, node attributes, and other known information \cite{lu2011link}. For example, LP is used to recommend friendships in social networks \cite{salamat2021heterographrec}, predict potential protein-protein interactions in drug discovery \cite{nasiri2021novel}, and enable targeted product recommendations on e-commerce platforms \cite{kaya2020hotel}. This highlights the practical importance of LP across multiple fields. However, existing link prediction methods face several important challenges. First, with the increasing scale of graph data, the efficiency and accuracy of existing methods in handling large-scale sparse graph data struggle to meet the demands of practical applications. Second, the issue of dataset mismatch means that models trained on one dataset often fail to adapt to other datasets, thus limiting the generalizability and applicability of link prediction techniques. How to effectively solve these problems and improve the performance of models in different scenarios is a critical challenge in the field of graph data mining. Overcoming these challenges will provide a theoretical foundation and technical support to promote graph data mining techniques in more practical applications.

Traditional network embedding methods, such as DeepWalk \cite{perozzi2014deepwalk} and Node2Vec \cite{grover2016node2vec}, have shown some success. However, these approaches face limitations when applied to large-scale datasets or complex topologies. These methods often lack scalability and effective parameter sharing. In contrast, graph neural networks (GNNs), which aggregate information from neighboring nodes to learn node embeddings \cite{wu2020comprehensive}, have emerged as a mainstream approach, providing better performance and efficiency. The graph convolutional network (GCN), introduced by Kipf and Welling \cite{kipf2016semi}, has been widely applied in LP tasks \cite{schlichtkrull2018modeling, chen2020multi}. However, GCNs primarily focus on aggregating information from local neighborhoods, making them less effective for capturing global relationships in large graphs. To address these limitations, graph attention networks (GATs) \cite{velivckovic2017graph} have been proposed. They use attention mechanisms to dynamically assign weights to neighbor nodes, offering greater flexibility in modeling long-range interactions and global structures \cite{ye2021sparse, wang2019heterogeneous}. Unlike GCNs, GATs are not restricted to fixed neighborhood structures. This ability to capture distant node interactions makes GATs particularly suitable for LP tasks in complex and large-scale graphs. This study builds upon GAT to develop an advanced LP framework.

As graph data continues to grow, models have become increasingly complex. To enhance the learning capability for large datasets, we incorporate a graph transformer. In traditional transformer architectures \cite{vaswani2017attention}, positional encoding represents the sequential relationships between nodes. However, such encodings struggle to effectively capture the topological relationships in graph structure data \cite{min2022transformer}. To address this, recent research has introduced improvements to the self-attention module \cite{ying2021transformers, wang2024automatic, li2022does}. The self-attention module is a powerful mechanism for learning node-pair relationships. In our study, we improve the graph transformer by embedding distant neighbor information as a bias in the attention matrix of the self-attention module. This helps the model better capture global relationships between nodes.

It is challenging to identify latent relationships between nodes, as graph data often suffers from sparsity and noise \cite{luo2023graph}. This "cold-start problem" hinders effective learning. While most existing methods focus on improving performance on single datasets, they overlook the potential of transfer learning to generalize across datasets. The pretrain–fine-tune is widely used in graph representation learning and has been developed for link prediction \cite{han2021adaptive}. Early work focused on alleviating graph scale gaps and data sparsity \cite{wang2021pre, yuan2023alex}, and later domain-adaptive methods aimed to reduce mismatches in scale, density, and structure between source and target graphs \cite{shao2024contrastive, shen2023domain}. However, existing approaches still face three related limitations when applied to cross-graph link prediction. First, topology/feature distribution mismatch: global priors learned on the source graph often struggle to directly adapt to the local patterns in the target graph, because most pretrained models lack explicit cross-domain adaptation mechanisms. Second, parameter and sample imbalance lead to overfitting and high cost: fine-tuning the entire large pretrained model on a small, sparse target graph is prone to overfitting and computationally expensive due to redundant trainable parameters. Third, distant structural information is difficult to transfer: conventional GNN or Transformer relies on local aggregation or positional encodings that do not robustly capture cross-domain long-range relations, limiting their ability to compensate for sparse edge information.

\begin{figure}[ht]
    \centering
    \includegraphics[clip,scale=0.33]{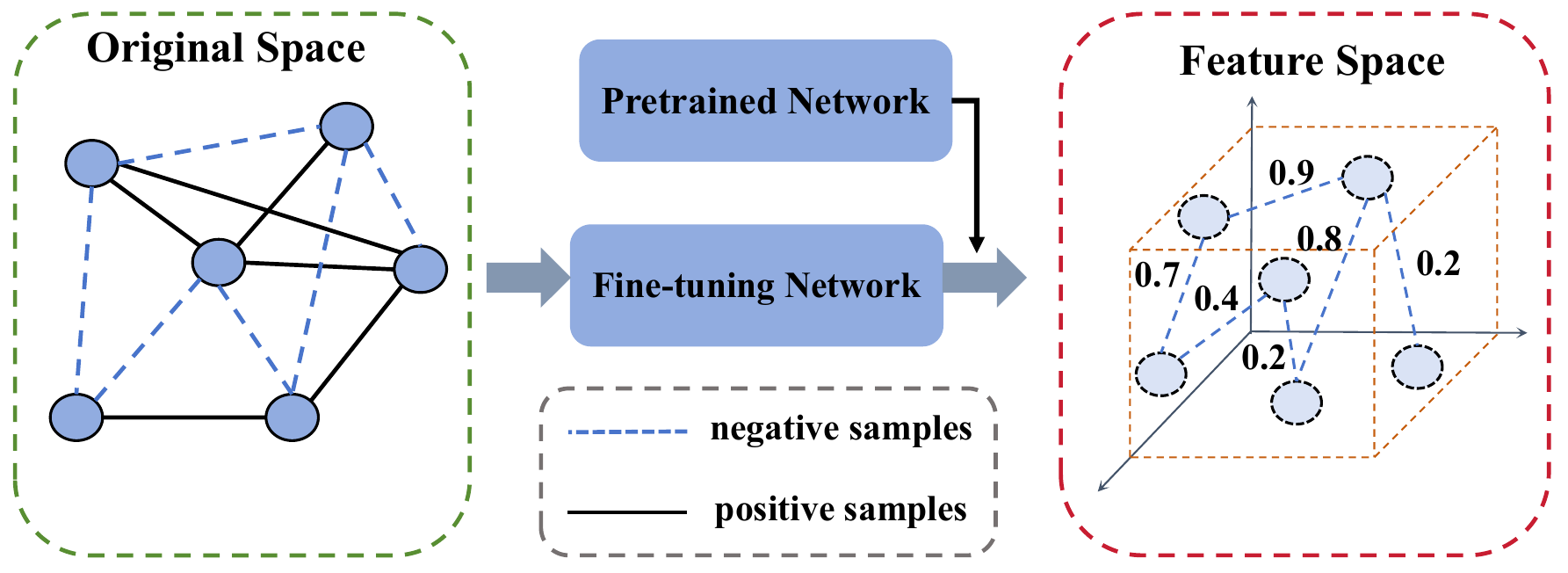}
    \caption{An overview of our transfer learning framework. Positive samples represent observed edges (\textit{label} = 1), while negative samples represent unobserved edges (\textit{label} = 0). In the feature space, a nonlinear method is used to compute $y_{\textit{pred}}$. If $y_{\textit{pred}} > \theta$, it indicates a potential edge formation. Here $\theta$ is a given threshold, typically set to 0.5.}
    \label{fig: link}
\end{figure}

We address these issues with three targeted designs: learn rich global structural priors during pretraining; during fine-tuning, freeze the backbone and replace the Transformer encoder layer with a lightweight self-adapter module while simplifying multi-head attention to single-head to greatly reduce trainable parameters and computation; and introduce distant-neighbor bias into self-attention to explicitly model long-range topology to enhance perception of sparse structures. In combination, these three components, structured pretraining, parameter-efficient adaptation, and long-range modeling, directly mitigate the main bottlenecks of the pretrain-fine-tune for cross-graph link prediction. Fig. \ref{fig: link} shows the details of the proposed transfer learning framework, called \underline{G}raph \underline{A}ttention \underline{A}daptive \underline{T}ransfer \underline{N}etwork (GAATNet), which combines pre-training and fine-tuning to adapt large-scale global information to smaller datasets while achieving efficient fine-tuning for LP tasks. The main contributions of this work are summarized as follows:

\begin{itemize}
    \item  We propose a pretraining and fine-tuning strategy to balance model complexity and dataset scale. This dual-stage learning effectively adapts to diverse graph sizes. 

    \item To address data sparsity and topology encoding challenges, we introduce diffusion-based data augmentation and embed distant neighbor information as bias in the attention module, enhancing the model’s ability to capture global structure.

    \item To improve fine-tuning efficiency, we add a self-adapter module, significantly reducing parameter complexity. This prevents overfitting and makes the model lightweight.

    \item We evaluate GAATNet on seven public datasets of varying scales. The results demonstrate that GAATNet consistently outperforms baseline methods in LP tasks. Additionally, the fine-tuning stage achieves faster training speeds due to the reduced parameter complexity, showing its combined performance and efficiency advantages. 
\end{itemize}

The remainder of this paper is structured as follows: Related work is reviewed and summarized in Section \ref{sec: relatedwork}. We provide a detailed explanation of the proposed GAATNet model in Section \ref{sec: Methodology}. Furthermore, we conducted some experiments, and the experimental results are presented in Section \ref{sec: experiments}. Finally, the conclusion and future work are discussed in Section \ref{sec: conclusion}.

\section{Related Work} \label{sec: relatedwork}

In this section, we provide a concise review and discussion of related work from three key perspectives: link prediction, graph transformer, and graph transfer learning.

\subsection{Link prediction}

Link prediction aims to predict future edges that may exist but have not been observed in graph structure data. Existing link prediction methods have achieved good results in many domains. These methods can be divided into two main types: embedding-based methods and GNN-based methods. However, with the increase in the scale and structural complexity of graph data, the existing methods still exhibit significant limitations in terms of computational efficiency, global information capturing ability, and model interpretability. Embedding-based methods represent nodes through random walk sequences, such as DeepWalk \cite{perozzi2014deepwalk} and node2vec \cite{grover2016node2vec}. They estimate the likelihood of a connection based on simple node pair information. Although these methods can effectively capture local relationships between nodes, they lack parameter-sharing mechanisms, have limited capability for modeling global information, and have poor scalability. They have difficulty meeting the task requirements of large-scale graphs.  

With advances in computing power, GNNs have been successfully applied to link prediction \cite{lu2011link}. It learns node embeddings through neighbor aggregation. GNN-based methods can be divided into subgraph-based, generative model-based, and explicit feature-based methods. Subgraph-based methods, such as SEAL \cite{zhang2018linkseal} and LGCL \cite{zhang2023linelgcl}, utilize local subgraph structures to capture information within a node's neighborhood. Although enclosed subgraph-based methods are more effective in modeling local relationships, their high computational costs and locality constraints make it difficult to generalize to larger datasets. For instance, SEAL emphasizes modeling of local relationships but has the high computational cost of extracting subgraphs, limiting its applicability to large-scale graphs. Similarly, LGCL improves subgraph extraction through contrastive learning on line graphs but still relies on local structures. Generative model-based methods, including VGAE \cite{kipf2016variational}, try to model the graph generation process. The generative model-based methods suffer from the same problems of high computational complexity and lack of interpretability \cite{kumar2020link}. In contrast, explicit feature-based methods, such as GAT \cite{velivckovic2017graph} and SIEG \cite{shi2024structural}, combine node attributes and graph topology information to extract effective node embeddings through GNNs from explicit neighborhood relationships and features. These methods consider more comprehensive information and obtain global embeddings with lower computational cost, exhibiting better performance. However, this approach still has room for improvement in handling large-scale sparse data and robustness to noise. 
Recent graph learning research has been applied to address graph sparsity and enhance stable generalization. For example, multi-topology contrastive graph representation learning was used to extract complementary information in sparse graphs \cite{xie2026multi}, while deep graph convolutional networks were used to analyze the key factors affecting generalization stability \cite{yang2025deeper}.

To address the limitations of existing methods in terms of computational efficiency, global information capture capability, and noise robustness, we propose the GAATNet framework. GAATNet combines GNNs and self-attention mechanisms, integrating the advantages of embedding-based methods for local information modeling and the global information capture capability of GNN-based methods. It reduces computational cost by migrating trained models into the fine-tuning network while also incorporating a diffusion data augmentation strategy to address the issue of data sparsity. Unlike existing methods, GAATNet maintains a low computational cost. It overcomes the limitations of local information capture, effectively captures the global structure of the graph, and shows improved performance on large-scale sparse graphs. GAATNet not only enhances the accuracy and generalization ability of link prediction but also provides new ideas and solutions.

\subsection{Graph transformer}

Transformers have shown impressive performance in natural language processing \cite{vaswani2017attention} and computer vision \cite{dosovitskiy2020image}. They are also increasingly being applied to graph data \cite{chen2022structure, kreuzer2021rethinking}. Graph Transformer combines GNNs and Transformer architectures. They use the Transformer’s representational power and flexibility to handle graph structure data effectively. However, there are two key challenges in applying Transformer to graph data: Data sparsity leads to model underfitting and the difficulty of using sequential positional encoding to represent the complex graph topology. For the first problem, graph data often has a small number of edges, many missing edges, and potentially noisy edges. To address this problem, Wu \textit{et al.} \cite{wu2023difformer} proposed a diffusion-based data augmentation strategy, which enhances the robustness of Transformer models in sparse graphs by augmenting graph data. It can reduce the impact of missing and noisy edges on embeddings. For the second problem, the traditional Transformer positional encoding is based on sequential order, which does not effectively capture the graph's topological structure. To solve this problem, Wang \textit{et al.} \cite{wang2024automatic} proposed Laplacian encoding, and Li \textit{et al.} \cite{li2022does} combined edge weight information with spatial encoding to represent relationships between nodes better. Although these methods improved the ability of the graph Transformer to model topological structures to some extent, they all overlooked the importance of neighboring nodes in representation learning. They do not consider the influence of neighboring nodes on the source node. Therefore, to avoid these problems in the GAATNet framework while taking neighbor relationships into account, we introduce a diffusion-based data augmentation method in the embedding matrix to enhance representation robustness in sparse graphs. Additionally, we integrate distant neighbor embedding information as a bias into the attention matrix in the \textit{Attention} layer. This enables GAATNet to better model long-distance dependencies and the influence of neighboring nodes. It provides more stable and accurate link prediction performance on large-scale sparse graph data.

\subsection{Graph transfer learning}

Graph transfer learning \cite{lee2017transfer} aims to transfer knowledge learned from one graph to another target graph, thus sharing information between different graph structure data. In recent years, self-supervised learning has achieved significant success in various graph downstream tasks \cite{you2020graph}. These advances have inspired interest in transferring learned graph embeddings to other graphs. However, cross-graph transfer learning faces several challenges \cite{han2021adaptive, wu2023non}. It mainly includes large differences in graph scale, feature sparsity, and noise interference, and limited labeled data in the target graph. When there is a significant scale difference between the source and target graphs, feature and topology mismatches often occur. Wang \textit{et al.} \cite{wang2021pre} proposed a cross-domain edge connection method, which alleviates the problem to some extent but still cannot fully eliminate the challenges of topology mismatch. In the case of sparse labeled data in the target graph, Yuan \textit{et al.} \cite{yuan2023alex} introduced graph contrastive learning and domain alignment strategies to transfer the knowledge of the source graph to the target graph. Although these methods improved transfer learning performance to some extent, they usually rely on complex model structures. Their high computational costs make it difficult to generalize them in practical large-scale applications.

Several graph domain adaptation methods have been proposed to reduce differences between source and target graphs \cite{wu2020unsupervised, you2023GraphDomainAda}. Dai \textit{et al.} \cite{dai2022graphAdaGCN} used adversarial domain adaptation to reduce distribution gaps and improve transfer. Shen \textit{et al.} \cite{shen2023domain} introduced a domain-adaptive graph attention supervised network for cross-network edge prediction. Shao \textit{et al.} \cite{shao2024contrastive} combined contrastive learning with selective self-training to enhance cross-domain node representation consistency. These works address scale, density, and structural mismatches from adversarial, alignment, and contrastive perspectives. However, existing domain adaptation methods often do not combine large pretraining with parameter-efficient fine-tuning. They incur high computational costs and tend to overfit or require many trainable parameters on small target graphs. To achieve more efficient cross-graph transfer learning in the GAATNet framework, we combine pre-training and fine-tuning strategies. In the pre-training stage, a complex model is used to capture global structural information from large-scale datasets. In the fine-tuning stage, a simpler model is employed to adapt to the local features of small-scale datasets. This approach balances global and local information, which helps alleviate data sparsity and feature mismatch. The GAATNet model better adapts to graphs of different scales, providing a robust and scalable transfer learning solution with efficient computation.

\section{Methodology} \label{sec: Methodology}

In this section, we introduce GAATNet, a novel framework for link prediction that integrates graph attention mechanisms with adaptive fine-tuning. We first define the link prediction problem and then describe the overall architecture and training strategy of GAATNet in detail. As shown in Fig. \ref{fig: GAATNet}, GAATNet consists of two main stages: pretraining and fine-tuning. Finally, we provide a comprehensive description of all its components and analyze the algorithm's complexity. The notations used in this paper are summarized in Table \ref{tb: notations}.

\begin{table}[ht]
\small
\centering
\caption{Notations}
\label{tb: notations}
\begin{tabular}{cl}
\toprule
\textbf{Notation} & \textbf{Description} \\
\midrule
$G$ = $(V, E, X, A)$ & An interaction network, $V$ is node set and $E$ is edge set  \\
$A$ & Adjacency matrix of $G$ \\
$X$ & Node embedding matrix of $G$ \\
$X_{init}$ & Initial node embedding matrix \\
$\vec{\mathbf{x}}_i$ & Embedding vector of node $v_i$ from $X$ \\
$Z$ & The output embedding matrix of the attention layer \\
$Z^{\prime}$ & The output embedding matrix of the Feature Transformer layer or Self-Adapter layer \\
$Z^{\prime\prime}$ & The final embedding matrix used for generating the prediction scores \\
$\vec{\mathbf{z}}_i$ & Embedding vector of node $v_i$ from $Z$ \\
$W$ & A learnable weight matrix \\
$\varphi_{dist}$ & Embedding information of the distant neighbor \\
$s_{ij}$ & The prediction scores \\
$\mathcal{L}_{link}, \mathcal{L}_{con}, \mathcal{L}_{total}$ & The link prediction loss, the contrastive loss, the total loss \\
$\lambda, \tau$ & A weighted hyperparameter and a temperature hyperparameter \\
\bottomrule
\end{tabular}
\end{table}

\subsection{Overall architecture of GAATNet}

\textbf{Problem definition}. As with most existing link prediction methods, we model a given interaction network as an undirected graph $G$ = $(V, E, X, A)$, where $V$ = \{$v_{1}$, $v_{2}$, $\dots$, $v_{n}$\} is a set of nodes with a size $n$, and $E$ = $\{e_{ij} | v_i \in V \cap v_j \in V\}$ is a set of edges with a size $m$. $A$ $\in \mathbb{R}^{|V| \times |V|}$ is the adjacency matrix of $G$. $A_{ij}$ = 1 if $e_{ij}$ $\in E$, and $A_{ij}$ = 0 otherwise. $X$ is the node embedding matrix, where each node $v \in V$ has a $d$-dimensional embedding vector $x_v \in \mathbb{R}^d$. Each edge is assigned a label $y$, where 1 indicates the presence of the edge in the graph, and 0 indicates its absence. Let $U$ = \{$u_1$, $u_2$, ..., $u_D$\} represent all possible link cases in the graph $G$, where $D$ = $\frac{n(n-1)}{2}$. The goal of link prediction is to find potential future links in the set $U-E$. Thus, it is a binary classification task for relationships between node pairs. For any two nodes $v_i$, $v_j$ $\in V$, the goal of our study is to design a model that learns the embedding representation Z between nodes. This is achieved by utilizing the node embedding matrix $X$ and the graph adjacency matrix $A$. It predicts the probability $P(e_{ij} | Z_i, Z_j)$ of a link existing between them.

\begin{figure*}[ht]
    \centering
    \includegraphics[clip,width=0.85\textwidth]{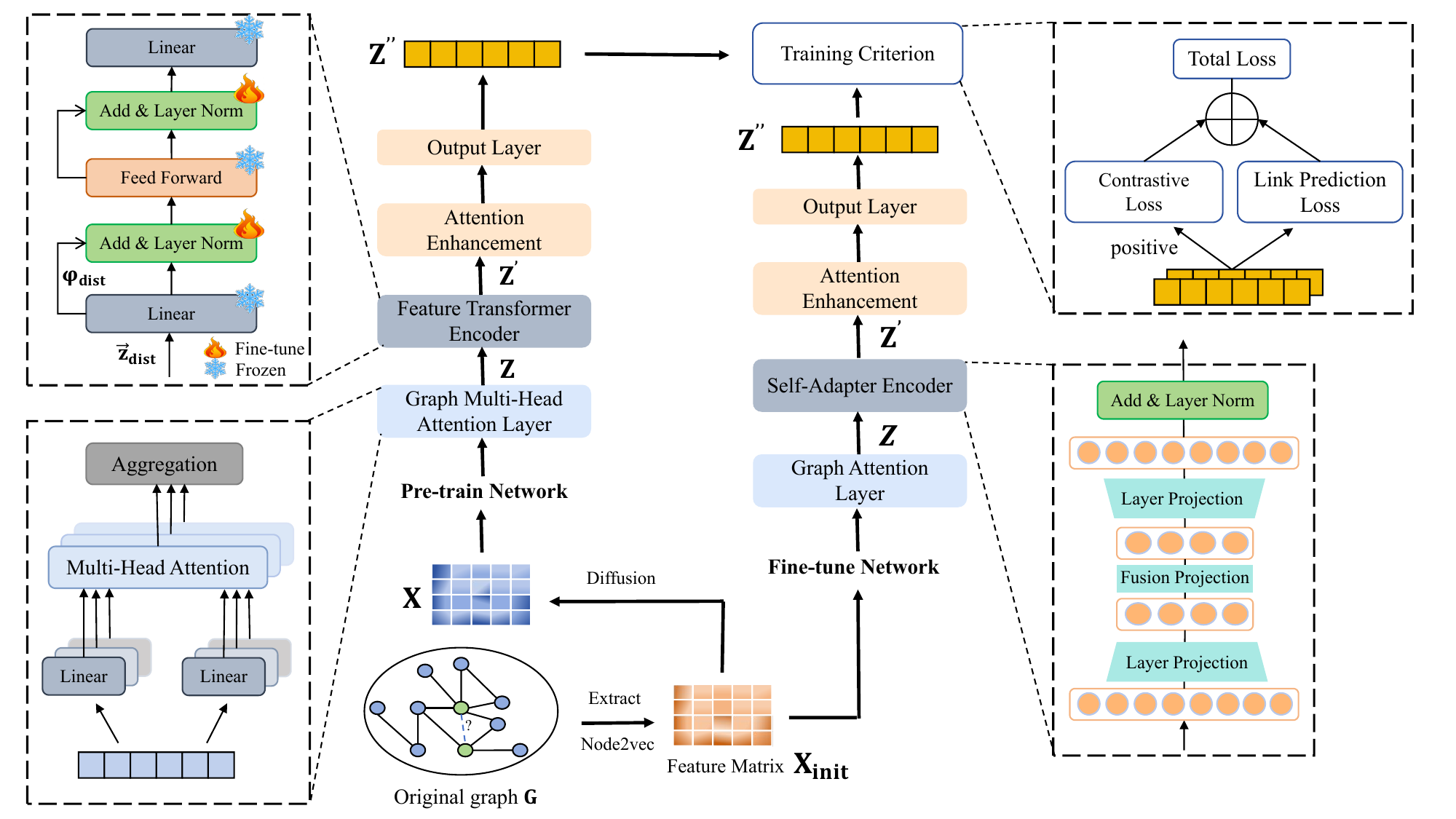}
    \caption{Overview of the GAATNet framework. The pre-training network (left) incorporates distant neighbor embeddings as attention biases. The fine-tuning network (right) uses a lightweight self-adapter module to boost efficiency and reduce trainable parameters. Both parts are trained under a joint objective combining contrastive loss and link prediction loss.}
    \label{fig: GAATNet}
\end{figure*}

We adopt a commonly used network embedding method to solve this binary classification task. Therefore, we propose GAATNet, a transfer learning framework based on graph attention networks, designed to enhance graph representation learning through adaptive mechanisms and two-stage training strategies. As shown in Fig. \ref{fig: GAATNet}, the framework integrates key components, including graph attention mechanisms, a feature transformation layer, and a lightweight self-adapter module, to achieve a balance between performance and computational efficiency. The input to the model is a graph $G$, where the graph's structural information is represented as an adjacency matrix $A$. The initial node embedding matrix $X \in \mathbb{R}^{n \times d}$ is learned by using the node2vec algorithm \cite{grover2016node2vec}. In the pretraining stage, $X$ undergoes a data augmentation process through diffusion to enrich the embedding space and promote effective information propagation, thereby smoothing the features of the entire graph structure. Subsequently, a graph multi-head attention layer dynamically computes the importance of neighboring nodes, learning a new embedding matrix that enables the model to focus on key links. The new embedding matrix enters the feature transformation layer to model global interactions between node embeddings. Following this, an attention-enhanced output layer extracts key link information and feeds it into the link prediction task. It then compares the features of labeled and known links to predict the probability of link existence. In the fine-tuning stage, to make the model lightweight, the initial node feature matrix is $X$ without the diffusion process. Additionally, a self-adapter layer replaces the feature transformation layer, which can efficiently learn the global interaction information between node embeddings without retraining the whole network.

\subsection{Data augmentation}

We first learn the initial node embedding matrix $X_{init}$ from the graph $G$ = $(V, E)$ by the Node2Vec algorithm \cite{grover2016node2vec}:
\begin{equation} \label{eq1}
    X_{\textit{init}} = \textit{Node2Vec}(G),
\end{equation}
where $X_{\textit{init}}$ $\in$ $\mathbb{R}^{n \times d_0}$, $n$ is the number of nodes, and $d_0$ is the initial embedding dimension. This algorithm learns meaningful representations solely from topological structures, making it applicable to datasets without raw node features. Additionally, it maps all datasets into a unified vector space through consistent output dimensions, thereby providing standardized inputs for subsequent attention layers and simplifying preprocessing. 

For data enhancement, the graph diffusion technique is introduced to smooth and augment the embedding matrix. Graph diffusion enriches the structural information of the graph, effectively alleviating the problem of data sparsity. Through the diffusion method, the features of noisy nodes are smoothed by their multi-hop neighboring nodes. This also prevents data sparsity from weakening the model's learning ability when the embedding enters the Transformer. This makes the model capture complex dependencies more effectively. This graph diffusion process is as follows \cite{nguyen2024diffusion}:
\begin{equation} \label{eq2}
    X^{(t+1)} = (1-\alpha) X_{init} + \alpha D^{-\frac{1}{2}} A D^{-\frac{1}{2}} X^{(t)},
\end{equation}
where $A$ is the adjacency matrix. $X^{(t)}$ denotes the embedding matrix generated at the t-step diffusion. $D$ is the degree matrix of the node, where $D_{ii}$ = $\sum_{j} A_{ij}$. $\alpha \in (0,1)$ is the diffusion proportion coefficient, controlling the weight of neighboring features. After $T$ iterations, the diffused embedding matrix $X^{T}$ is obtained. We will refer to $X^{T}$ as $X$ in the following discussion for simplicity.

\subsection{Pre-training network}

\begin{figure}[ht]
    \centering
    \includegraphics[clip,scale=0.35]{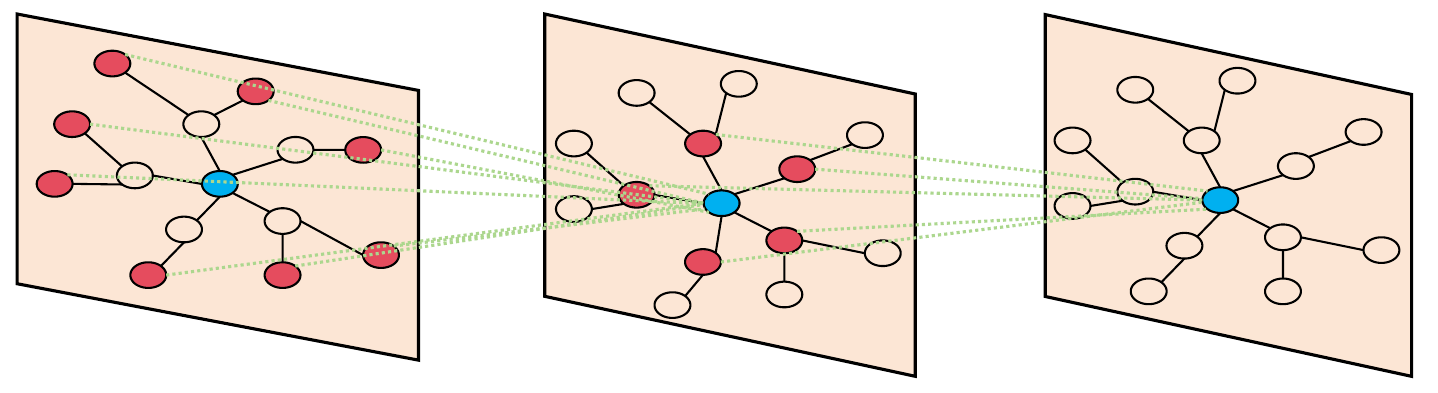}
    \caption{The process of aggregating node features from neighboring information in GAT.}
    \label{fig: GAT}
\end{figure}

The pre-training phase consists of two main components: the graph multi-head attention layer and the feature transformer encoder layer. The input to the graph multi-head attention layer is the embedding $\vec{\mathbf{x}}_v \in \mathbb{R}^d$ of each node. The first step is to convert the input embeddings into an attention correlation coefficient, as shown in the following formula \cite{velivckovic2017graph}:
\begin{equation} \label{eq3}
    e_{ij}^{(k)} = \text{LeakyReLU}(\vec{a}^T[W^{(k)} \vec{\mathbf{x}}_i \| W^{(k)} \vec{\mathbf{x}}_j]).
\end{equation}
Here $e_{ij}$ is the attention coefficient obtained from the attention calculation, indicating the degree of importance of node $v_j$ to node $v_i$. \textit{LeakyReLU} is an activation function and $k$ is the $k$-th attention head. $\vec{\mathbf{x}}_i$ and $\vec{\mathbf{x}}_j$ are the embeddings of adjacent nodes $v_i$ and $v_j$, respectively. $W \in \mathbb{R}^{d \times d^{\prime}}$ is a learnable weight matrix, where $d$ and $d^{\prime}$ represent the input and output mapping dimensions, respectively. $\|$ denotes the concatenation operation. $\vec{a} \in \mathbb{R}^{2d^{\prime} \times 1}$ is a learnable attention vector. $e_{ij}$ is normalized to obtain the attention weight \cite{velivckovic2017graph}:
\begin{equation} \label{eq4}
    \alpha_{ij}^{(k)} = \text{softmax}_j(e_{ij}^{(k)}) = \frac{\exp(e_{ij}^{(k)})}{\sum_{v_r \in \mathcal{N}(v_i)} \exp(e_{ir}^{(k)})},
\end{equation}
where $\alpha_{ij}$ is the final importance ratio of the node $v_j$ to the node $v_i$. $\mathcal{N}(v_i)$ is the set of neighboring nodes of node $v_i$. This calculation ensures that the attention weights for all neighbors of the node $v_i$ sum to 1. Therefore, we obtain a single-head attention layer to update the new embedding representation vector of a node $v_i$ in the $k$-th head, as shown in the following formula \cite{velivckovic2017graph}:
\begin{equation} \label{eq5}
    \vec{\mathbf{x}}^{\prime(k)}_i = \sigma \left(\sum_{v_j \in \mathcal{N}(v_i)} \alpha_{ij}^{(k)} \mathbf{W}^{(k)}\vec{\mathbf{x}}^{(k)}_j\right),
\end{equation}
where $\sigma(\cdot)$ is an ELU activation function, enhancing the model's nonlinear representation capability. To enhance the learning capability for node embeddings, we use the graph multi-head attention mechanism. By concatenating multiple single-head attention layers, we compute the new node embedding vectors. As follows \cite{velivckovic2017graph}:
\begin{equation} \label{eq6}
    \vec{\mathbf{z}}_i = \underset{k=1}{\overset{K}{\|}} \vec{\mathbf{x}}^{\prime(k)}_i,
\end{equation}
where $K$ is the number of attention heads. In summary, the node embedding matrix updated through the graph multi-head attention layer is denoted as $Z$ = $(\vec{\mathbf{z}}_1$, $\vec{\mathbf{z}}_2$, ..., $\vec{\mathbf{z}}_n)^T$, $\vec{\mathbf{z}}_i \in \mathbb{R}^{1 \times d^{\prime}}$. During this process, the node embeddings become more concentrated and distinct. The process of node aggregation through GAT is shown in Fig. \ref{fig: GAT}.

In the graph multi-head attention layer, embeddings become more focused and prominent, but this may result in the loss of some structural information within the graph. The transformer also struggles to use sequential positional encoding to represent the complex topology of graphs \cite{min2022transformer}. To enable the model to learn more accurate embeddings and further capture global features, we adjust the original transformer structure to include some embedding information of the distant neighbor nodes, specifically embeddings of some $g$-hop($g \geq 2$) neighbors. This adjustment is motivated by the fact that graph structures often contain valuable global context that is not always captured by local node interactions alone. By introducing distant neighbors' embeddings into the self-attention mechanism, we allow the model to retain a broader view of the graph structure, thus improving its ability to generalize and make predictions. First, we randomly extract some $g$-hop neighbors using the adjacency matrix and then use their embeddings $\vec{\mathbf{z}}_{dist}$ to the linear layer to encode as follows:
\begin{equation} \label{eq7}
    \varphi_{dist} = \text{Linear}(\vec{\mathbf{z}}_{dist}),
\end{equation}
where $\varphi_{dist}$ is the output with the embedding information of the distant neighbor. Then, this output is used as a bias term in the self-attention mechanism \cite{vaswani2017attention}, allowing it to participate in the attention score matrix of the transformer, as shown in the following equation:
\begin{equation} \label{eq8}
    \text{SelfAttention}(Q, K, V) = 
    \text{softmax}\left(\frac{QK^T}{\sqrt{d_k}} + \varphi_{\textit{dist}}\right)V,
\end{equation}
where $Q$ = $ZW_Q$, $K$ = $ZW_K$, and $V$ = $ZW_V$ represent queries, keys, and values, respectively. $W_Q$, $W_K$, and $W_V$ are the learnable weight matrix. $d_k$ is the dimension of the key vector. 

The novel idea behind using distant neighbor embeddings as a bias in the attention mechanism is to enhance the model's ability to capture global features that are critical for link prediction. Traditional attention mechanisms often focus on local neighborhood interactions, but on large sparse graphs, this may miss crucial long-range dependencies between distant nodes. By introducing "distant neighbors" as a bias term, we provide a mechanism for the model to incorporate a broader context when computing attention scores. This improves the overall quality of the learned embeddings and facilitates the capture of more global structural features. This strategy helps mitigate the limitations of local attention and enhances the model's robustness in link prediction, particularly when dealing with large-scale data.

Negative transfer occurs when knowledge from the source domain harms performance on the target domain, for example, by introducing source-specific features or noisy signals. The attention mechanism in GAATNet mitigates this issue via adaptive re-weighting. It assigns higher weights to features that are informative across domains and lower weights to source-specific or noisy signals. This selective weighting reduces the contribution of mismatched information, thereby limiting the representation gap between the source and target. In effect, attention works like importance weighting in domain adaptation, learning which information is transferable. Combined with our pretrain–finetune, the attention bias enhances globally consistent patterns and suppresses local noise, thus reducing the risk of negative transfer.

The latter steps are the same as in the original transformer module. After passing through the feedforward neural network (FFN) and layer normalization (LayerNorm), the final new node embedding matrix $Z^{\prime} \in \mathbb{R}^{n \times d^{\prime}}$ is obtained. This process helps in information compression and global feature extraction. Finally, we use Equation (\ref{eq5}) to obtain the attention-enhanced output, which also achieves the purpose of feature dimension reduction. This facilitates the computation of the subsequent link prediction and generates the final low-dimension embedding $Z^{\prime\prime} \in \mathbb{R}^{n \times d^{\prime\prime}}$, where $d^{\prime\prime}$ represents the output's reduced dimension.

\subsection{Fine-tuning network}

To retain the valuable node embedding information from the pre-trained network, we freeze the majority of the parameters from the pre-training phase and focus solely on training the parameters of the layer normalization layers. This approach allows the pre-trained network to more effectively adapt to the node embeddings in the fine-tuning network. Thus, the fine-tuning network can efficiently leverage the existing node embeddings, avoiding the need for training from the beginning, and ultimately leading to improved performance and faster convergence.

At this stage, the node embedding matrix is no longer processed by diffusion to make the model more lightweight. Instead, it is directly fed into a graph single-head attention network, where the computation follows Equation (\ref{eq5}). This output $Z \in \mathbb{R}^{n \times d^{\prime}}$ is then used as the input to the self-adapter encoder layer, which enhances the embedding representation capability and accelerates the learning speed of the model.

In the fine-tuning network, we designed a parallel self-adapter structure that is inserted between the graph attention layer and the attention-enhancement layer. During adapter-based fine-tuning, only a small number of trainable parameters are introduced, while a residual structure is used in the model to learn key features. 

Specifically, we use layer projection and fusion projection neural networks as the main adapter components. The adapter module tunes embedding through a bottleneck layer. The projection neural network first maps the embedding matrix from $d$-dimension to $q$-dimension, then continues a dimension-invariant projection in $q$-dimension to fuse key features, and finally maps them back to $d$-dimension. This low-dimensional representation allows the model to focus on the most important features, reducing the computational burden. The output of the adapter uses a residual structure, where the projected information is aggregated with the original information as the final output. The key optimization of the self-adapter module lies in the efficient tuning of parameters during the fine-tuning process. By introducing only a small number of trainable parameters, the self-adapter enables the model to focus on optimizing key features without overfitting the training data. This optimization significantly improves the training efficiency, as the model is able to adapt to the task with minimal adjustment to the existing embeddings. Moreover, the residual connection in the self-adapter prevents the network from losing the original pre-trained information, thus improving stability and mitigating overfitting.
This output matrix $Z^{\prime}$ is computed as follows:
\begin{equation} \label{eq9}
    Z^{\prime} = Z + \text{GELU} \Bigl[\text{GELU}(Z W_1) W_2\Bigr] W_3,
\end{equation}
where $W_1$ $\in $ $\mathbb{R}^{d' \times q}$ and $W_3$ $\in $ $\mathbb{R}^{q \times d'}$ are layer projection weight matrix, and $W_2$ $\in$ $\mathbb{R}^{q \times q}$ is the fusion projection weight matrix. $\text{GELU}(\cdot)$ is a nonlinear activation function. We set $q < < d'$, introducing only a small number of trainable parameters while retaining the key embedding information for each node. This lightweight self-adapter module helps balance model complexity and performance by maintaining a small parameter set while ensuring the preservation and enhancement of node embeddings. It contributes to more efficient fine-tuning and accelerates convergence, thereby improving the overall robustness of the model.

Similarly, we use Equation (\ref{eq5}) to obtain the attention enhancement output, which is the same process as in the pretraining network. Here, both the pretraining network and the fine-tuning network generate a new embedding matrix $Z^{\prime\prime}$ $\in$ $\mathbb{R}^{n \times d^{\prime\prime}}$, where $d^{\prime\prime}$ represents the output dimension of the attention enhancement layer. This embedding is subsequently used for the link prediction task to compute the total loss.

Conventional transfer learning either fine-tunes the entire model or only the top classifier. Traditional methods typically treat a pretrained model as a feature extractor and adapt to a new task by fine-tuning its output layer, which can still require retraining many parameters. By contrast, GAATNet has undergone key structural modifications while retaining the frozen parameters of the pretrained model backbone: (1) replacing the feature transformer encoder layer with a lightweight self-adapter module; and (2) simplifying multi-head attention to a single-head attention mechanism to further reduce computational complexity. Additionally, only the layer normalization parameters are fine-tuned. This strategy significantly reduces the number of trainable parameters, thereby speeding up fine-tuning and alleviating overfitting on small datasets, and enabling more efficient and robust knowledge transfer.

\subsection{Link prediction}

We determine the likelihood of a link existing in the graph by estimating the probability of its presence. This is achieved by calculating the embedding score of the corresponding edge nodes. Non-linear methods are often more effective than linear approaches (e.g., directly computing the vector dot product) \cite{kumar2020link}. In this paper, we use the path Euclidean distance as a non-linear method to calculate the similarity score between node pairs. A higher similarity score indicates a more likely link between the two nodes. The embedding vectors $(\vec{\mathbf{z}}_i, \vec{\mathbf{z}}_j)$ of two edge nodes $(v_i, v_j)$ are taken as inputs to obtain the score $s_{ij}$ of the corresponding link. It is calculated as follows:
\begin{equation} \label{eq10}
    s_{ij} = \exp\Bigl\{-\text{ReLU}\Bigl[\psi \|\vec{\mathbf{z}}_i - \vec{\mathbf{z}}_j\|^2\Bigr] \Bigr\}, 
\end{equation}
where $\text{ReLU}(\cdot)$ is an activation function. $\psi$ is a learnable parameter matrix. The similarity score $s_{ij}$ obtained through Equation (\ref{eq10}) is a value in the range $(0, 1]$. Closer node pairs will have higher scores, as the Euclidean distance between the node pairs tends to 0, while pairs of nodes farther away will have lower scores. Therefore, we can use $s_{ij}$ to calculate the link prediction loss, as shown in the following formula:
\begin{equation} \label{eq11}
    \small
    \mathcal{L}_{link} = -\frac{1}{|n_{total}|}\sum_{(i,j) \in E_{total}} \Bigl[y_{ij}\log s_{ij} + (1-y_{ij})\log(1-s_{ij})\Bigr], 
\end{equation}
where $n_{total}$ = $n_{pos}$ + $n_{neg}$, with $n_{pos}$ and $n_{neg}$ representing the number of edges in the existing link set $E$ and the constructed negative sample link set, respectively. The values of $n_{pos}$ and $n_{neg}$ are equal. $y_{ij}$ is the true label of the edge $e_{ij}$. $E_{total}$ is the sum of the set of edges $E$ and negative samples.

To enhance the model's robustness to noisy data, we employ the InfoNCE contrastive loss \cite{oord2018representation}. The objective is to minimize the distance between node pairs with existing links and reduce the similarity of negative sample node pairs. This helps in learning an embedding that distinguishes between different node pairs. The formula for calculating the contrastive loss is as follows:
\begin{equation} \label{eq12}
    \mathcal{L}_{con} = -\frac{1}{|n_{pos}|}\sum_{(i,j) \in E} \log\frac{\text{exp}(\text{sim}(\vec{\mathbf{z}}_i, \vec{\mathbf{z}}_j)/\tau)}{\sum_{k \in E_{total}} \text{exp}(\text{sim}(\vec{\mathbf{z}}_i, \vec{\mathbf{z}}_k)/\tau)},
\end{equation}
\begin{equation} \label{eq13}
    \text{sim}(\vec{\mathbf{z}}_i, \vec{\mathbf{z}}_j) = \frac{\vec{\mathbf{z}}_i^T \vec{\mathbf{z}}_j}{\|\vec{\mathbf{z}}_i\| \|\vec{\mathbf{z}}_j\|},
\end{equation}
where $sim(\cdot, \cdot)$ is cosine similarity. $\tau$ is the temperature hyperparameter, controlling distribution smoothness. Then, the total loss of the GAATNet model is shown as follows:
\begin{equation} \label{eq14}
    \mathcal{L}_{total} = \mathcal{L}_{link} + \lambda \mathcal{L}_{con}, 
\end{equation}
where $\lambda$ is a weighted hyperparameter that balances the link prediction and the contrastive loss.

\subsection{Algorithm and complexity analysis}

To theoretically analyze the GAATNet model training efficiency and inference time complexity, we summarize the overall process of GAATNet in Algorithm \ref{alg: GAATNet}. From Algorithm \ref{alg: GAATNet}, we can analyze the time complexity in two parts: the pre-trained network and the fine-tuned network.

\begin{algorithm}
\footnotesize
\caption{The training process of GAATNet}
\label{alg: GAATNet}
\begin{algorithmic}[1]
\renewcommand{\algorithmicrequire}{\textbf{Input:}}
\renewcommand{\algorithmicensure}{\textbf{Output:}}
\REQUIRE{Original graph $G$ = $(V,E,X,A)$; Edge set $E_{pos}$; and Maximum iterations MaxIter.}
\ENSURE{Total loss $\mathcal{L}_{\textit{total}}$ and similarity score $s_{ij}$.}
\
\STATE{The initial features $X_{\textit{init}}$ performs diffusion to obtain $X$ via Eq. (\ref{eq2})}
\STATE{Randomly sample nonexistent edges to form edge set $E_{neg}$, and $E_{\textit{train}}$ = $E_{pos}$ + $E_{neg}$}  \\
\textbf{// Pretraining network}
\FOR{epoch = 1...MaxIter}
    \STATE{Calculate attention correlation coefficient $e_{ij}$ and normalization to obtain $\alpha_{ij}$ via Eq. (\ref{eq3}) and Eq. (\ref{eq4})}
    
    \STATE{Calculate the $k$-th new embedding via Eq. (\ref{eq5})\\ 
    Get new embedding matrix $Z$ via Eq. (\ref{eq6})}
    
    \STATE{Get $\vec{\mathbf{z}}_{\textit{dist}}$ from $g$-hop neighbor nodes.\\ Calculate $\varphi_{\textit{dist}}$ via Eq. (\ref{eq7}), update $\text{Attention}$ via Eq. (\ref{eq8}).\\
    Get $Z^{\prime}$ by Transformer}

    \STATE{Get attention enhancement output $Z^{\prime\prime}$ via Eq. (\ref{eq5})}

    \STATE{Calculate similarity score $s_{ij}$ via Eq. (\ref{eq10})}

    \STATE{Calculate link prediction loss $\mathcal{L}_{\textit{link}}$ and contrastive loss $\mathcal{L}_{con}$ via Eq. (\ref{eq11}) and Eq. (\ref{eq12})}

    \STATE{Calculate total loss via Eq. (\ref{eq14})}

    \STATE{Update pretraining parameters and save the best model}
\ENDFOR     \\
\textbf{// Fine-tuning network}
\STATE{Load pre-trained model. Freeze pre-trained parameters and update only parameters of \text{Norm}}
\FOR{epoch = 1...MaxIter}
    \STATE{Calculate attention correlation coefficient $e_{ij}$ and normalization to obtain $\alpha_{ij}$ via Eq. (\ref{eq3}) and Eq. (\ref{eq4})}
    
    \STATE{Get the new embedding matrix $Z$ via Eq. (\ref{eq5})}
    
    \STATE{Get $Z^{\prime}$ in adapter layer via Eq. (\ref{eq9})}

    \STATE{Get attention enhancement output $Z^{\prime\prime}$ via Eq. (\ref{eq5})}

    \STATE{Calculate similarity score $s_{ij}$ via Eq. (\ref{eq10})}

    \STATE{Calculate link prediction loss $\mathcal{L}_{\textit{link}}$ and contrastive loss $\mathcal{L}_{con}$ via Eq. (\ref{eq11}) and Eq. (\ref{eq12})}

    \STATE{Calculate total loss $\mathcal{L}_{\textit{total}}$ via Eq. (\ref{eq14})}

    \STATE{Update fine-tuning network parameters}
\ENDFOR
\RETURN {Total loss $\mathcal{L}_{\textit{total}}$ and similarity score $s_{ij}$}
\end{algorithmic}
\end{algorithm}

The pre-trained network mainly includes data augmentation, graph multi-head attention, feature transformer, and attention enhancement output. The time complexity of data augmentation is $O(mT)$, where $T$ represents the number of diffusion steps and $m$ is the number of edges. The graph multi-head attention mechanism computes attention coefficients for all edges. The complexity per head is $O(md+nd')$, where $d$ and $d'$ are the input and output dimensions, respectively, and $n$ is the number of nodes. For $K$ heads, the total time complexity is $O(K(md+nd'))$. The feature transformer layer introduces additional costs in the $\text{SelfAttention}$ layer for calculating attention scores, with a time complexity of $O(n^2d'+n(d')^2)$. The feedforward network and LayerNorm layer have a combined complexity of $O(2n(d')^2)$. Thus, this results in a total complexity of $O(n^2d'+3n(d')^2)$. The attention enhancement layer has a time complexity of $O(md'+nd'')$. For the loss computation, calculating the link prediction loss and contrastive loss has a combined time complexity of $O(2md'')$. In summary, the total time complexity of the pretraining network is $O(m(T+d'+Kd+2d'')$ + $n(Kd'+d''+3(d')^2)$ + $n^2d')$.

The fine-tuning network mainly includes graph single-head attention, the self-adapter module, and the attention enhancement layer. The time complexity of the graph single-head attention is $O(md+nd')$. The self-adapter module is transformed into a lightweight residual mapping with a time complexity of $O(2nd'q+nq^2)$, where $q$ is the mapping dimension. The subsequent computations have the same time complexity as those in the pretraining network. Therefore, the total time complexity is $O(m(d+d'+2d'')$ + $n(d'+2d'q+q^2+d''))$.

The fine-tuning network employs the self-adapter module to significantly reduce the number of trainable parameters, from $O(d'^2)$ to $O(d'q)$, where $q << d'$ is set. This ensures that the fine-tuning network remains lightweight and maintains satisfactory performance.

\section{Experiments} \label{sec: experiments}

In this section, we conduct extensive experiments to evaluate the effectiveness of GAATNet on seven public datasets and analyze several key aspects of the model. Our experiments were carried out on a system equipped with the Nvidia GeForce RTX 4060 Ti GPU, with 32 GB RAM. All experiments were implemented based on Python 3.8. The source code and datasets are publicly available at \url{https://github.com/DSI-Lab1/GAATNet}.

\subsection{Experiment setup}

\textbf{Datasets}. We conducted extensive experiments on seven public graph datasets, and the statistics of these datasets are shown in Table \ref{table: datasets}. Cora and Citeseer \cite{sen2008collective} are two citation networks, where each node represents a paper, and the edges represent citation relationships between papers. Facebook\footnote{\label{foot}http://snap.stanford.edu} is a social network sourced from the Facebook website, where each node is a user, and edges represent follower relationships. ChCh-Miner\textsuperscript{\ref{foot}} is a biomedical network, where nodes represent drugs, and edges represent drug interactions. Coauthor-CS \cite{shchur2018pitfalls} is a co-authorship network, where nodes are authors of a paper, and edges represent collaborative relationships between authors. Computers \cite{shchur2018pitfalls} is a co-purchase network, where nodes represent computer-related products, and edges represent two products being purchased by users at the same time. Email-Enron\textsuperscript{\ref{foot}} is a communication network, where nodes represent users, and edges indicate email communication between two people. We classify Cora and Citeseer as small datasets, Facebook and ChCh-Miner as medium datasets, and the remaining as large datasets, based on the number of edges. These datasets have different characteristics and allow a comprehensive evaluation of the proposed method.

\begin{table}[ht] 
     \footnotesize
     \centering
     \caption{Statistical overview of the datasets. $|\textit{Nodes}|$ = $n$ denotes the number of nodes, and $|\textit{Edges}|$ = $m$ denotes the number of edges. These edges are all existing edges and are positive samples. \textit{Features} denotes the dimension of the node embedding vectors. \textit{Density} = $\frac{2 m \times 10^3}{n(n - 1)}$ is the density of the graph.}
    \begin{tabular}{>{\centering\arraybackslash}p{2.5cm}>{\centering\arraybackslash}p{2.5cm}>{\centering\arraybackslash}p{1.2cm}>{\centering\arraybackslash}p{1.2cm}>{\centering\arraybackslash}p{1.2cm}>{\centering\arraybackslash}p{1.2cm}}
    \hline
         \textbf{Dataset} & \textit{Type} & \textit{Nodes} & \textit{Edges} & \textit{Features} & \textit{Density} \\
    \hline
         Cora & Citation & 2708 & 5429 & 256 & 1.481\\ 
         Citeseer & Citation & 3312 & 4732 & 256 & 0.863\\ 
         Facebook & Social & 4039 & 88234 & 256 & 10.820\\ 
         ChCh-Miner & Biomedical & 1514 & 48514 & 256 & 42.358\\ 
         Coauthor-CS & Co-authorship & 18333 & 163788 & 256 & 0.975\\ 
         Computers & Co-purchase & 13752 & 491722 & 256 & 5.201\\ 
         Email-Enron & Communication & 15642 & 137022 & 256 & 1.120\\ 
    \hline
    \end{tabular}
    \label{table: datasets}
\end{table}

\textbf{Baselines and evaluation metrics}. We compared GAATNet with a range of different baseline methods. (1) Classical link prediction methods such as GAT \cite{velivckovic2017graph} and GCN \cite{kipf2016semi}. (2) Link prediction methods based on subgraph or methods that combine explicit features with graph neural networks, such as SEAL \cite{zhang2018linkseal}, SIEG \cite{shi2024structural}, and HL-GNN \cite{zhang2024heuristic}. (3) Methods based on transfer learning/graph domain adaptation or on contrastive learning, such as AdaGCN \cite{dai2022graphAdaGCN}, LGCL \cite{zhang2023linelgcl}, LPFormer \cite{shomer2024lpformer}, and Graph2Feat \cite{samy2023graph2feat}. AdaGCN and LPFormer are related to graph domain adaptation; Graph2Feat primarily uses knowledge distillation; LGCL is associated with contrastive learning. This comprehensive comparison allows us to evaluate the proposed GAATNet's performance and accuracy. For evaluation, we use AUC and the F1 score as performance evaluation metrics. To reduce experimental errors, we conducted the experiments five times and took the average as the final result.

\textbf{Experimental settings}. For each graph structure dataset, the set of edges is split into train, validation, and test sets in an 8:1:1 ratio. These edges are positive samples. The positive samples in the test set are not visible during the training process. To ensure that the node embedding vectors contain richer information, we use the node2vec model to generate initial node embeddings with the dimension of the features set to 256. For our proposed GAATNet model, in the pretraining stage, the number of diffusion steps is set to 50. The first layer consists of a GAT with 4 heads. The second layer adopts a graph transformer structure, using 2 transformer encoder layers with an intermediate dimension of 64 and 4 attention heads. For extracting distant neighbor nodes, we set $g$ = \{2, 3\}. The third layer uses a single-head GAT. In the fine-tuning stage, the first layer uses a single-head GAT, while the second layer employs a self-adapter structure with a projection dimension of 8. During training, we use a learning rate of 0.01 and Adam as the optimizer, along with an early stopping strategy. To prevent overfitting, we apply a dropout rate of 0.3 and L2 regularization with a coefficient of 0.0005.

To include single-network baselines in the transfer learning comparison, we unified the experimental setup. All baselines use 256-dim embeddings from node2vec as input and share the same data split, negative sampling, and training criterion to ensure fairness. For non-transfer learning baselines (GCN, GAT, SEAL, SIEG, HL-GNN, LGCL, and LPFormer), we run two modes and report the best result as the final value: (1) no-transfer, trained from scratch on the target graph; (2) pretrain–finetune, pretrained on a large source graph and then transferred to the target graph with only the output layer finetuned (i.e., the standard transfer strategy). For methods that already include transfer learning (AdaGCN and Graph2Feat), we reproduce the migration procedures described in their original papers. To keep model capacity and computational cost comparable, we align hyperparameters such as hidden dimensions, attention heads, and layer counts with GAATNet, and perform dimension alignment when input/output sizes differ.

\subsection{Comparison with baselines}

\textbf{Comparison under balanced settings}. For this experiment, we sampled negative edges from the unobserved set equal in number to the positive edges, maintaining a 1:1 ratio. In our experiments, to ensure a fair comparison, we set the training epoch to 200 and conducted several experiments to obtain more robust results, and finally took the average value as the final experimental results. For our GAATNet, we used the largest dataset (Amazon Computers) as the source graph for pre-training. The experimental results are shown in Table \ref{tab: results}. It can be observed that GAATNet consistently outperforms all baseline methods on all datasets in both AUC and F1 score. This demonstrates its effectiveness and robustness in link prediction.

\begin{table*}[t]
\centering
\caption{Results (\%) on link prediction benchmarks after training for 200 epochs on seven datasets. \textbf{Bold} numbers denote the best performances, and \underline{underline} numbers denote the second-best performances.}
\resizebox{\textwidth}{!}{
\begin{tabular}{lcccccccccccccc}
\toprule
& \multicolumn{2}{c}{Cora} & \multicolumn{2}{c}{Citeseer} & \multicolumn{2}{c}{Facebook} & \multicolumn{2}{c}{ChCh-Miner} & \multicolumn{2}{c}{Coauthor-CS} & \multicolumn{2}{c}{Amazon Computers} & \multicolumn{2}{c}{Email-Enron}\\
\cmidrule(lr){2-3} \cmidrule(lr){4-5} \cmidrule(lr){6-7} \cmidrule(lr){8-9} \cmidrule(lr){10-11} \cmidrule(lr){12-13} \cmidrule(lr){14-15}
& \multicolumn{1}{c}{AUC} & \multicolumn{1}{c}{F1-Score} 
& \multicolumn{1}{c}{AUC} & \multicolumn{1}{c}{F1-Score} 
& \multicolumn{1}{c}{AUC} & \multicolumn{1}{c}{F1-Score} 
& \multicolumn{1}{c}{AUC} & \multicolumn{1}{c}{F1-Score} 
& \multicolumn{1}{c}{AUC} & \multicolumn{1}{c}{F1-Score} 
& \multicolumn{1}{c}{AUC} & \multicolumn{1}{c}{F1-Score} 
& \multicolumn{1}{c}{AUC} & \multicolumn{1}{c}{F1-Score}\\
\midrule
GCN  & $68.36\pm1.09$ & $58.23\pm1.73$ & $69.13\pm0.94$ & $60.26\pm1.78$ 
     & $65.37\pm2.93$ & $59.35\pm2.17$ & $66.05\pm1.32$ & $59.57\pm0.77$ 
     & $64.79\pm0.64$ & $58.21\pm1.12$ & $61.36\pm0.72$ & $55.38\pm0.44$ 
     & $65.23\pm1.93$ & $60.47\pm0.68$\\
GAT  & $83.78\pm0.66$ & $78.42\pm1.44$ & $85.94\pm0.89$ & $80.07\pm0.83$ 
     & $82.18\pm1.31$ & $77.53\pm0.93$ & $83.01\pm0.65$ & $79.84\pm0.77$
     & $80.66\pm0.54$ & $75.41\pm1.39$ & $82.78\pm0.76$ & $73.37\pm0.83$ 
     & $79.93\pm2.04$ & $74.15\pm0.33$\\
\midrule
SEAL   & $88.71\pm0.96$ & $81.34\pm1.41$ & $90.89\pm0.77$ & $82.81\pm0.81$ 
       & $86.47\pm0.83$ & $80.45\pm1.05$ & $88.56\pm0.15$ & $82.44\pm0.23$ 
       & $85.32\pm0.38$ & $80.71\pm0.19$ & $83.87\pm0.93$ & $79.33\pm1.21$ 
       & $84.49\pm0.64$ & $78.59\pm0.21$\\
SIEG   & $91.45\pm1.03$ & $83.54\pm0.44$ & $92.10\pm0.21$ & $84.01\pm0.54$ 
       & $90.72\pm0.51$ & $84.41\pm0.42$ & $87.44\pm0.13$ 
       & $\underline{82.93\pm0.37}$ & $88.11\pm2.05$ & $82.30\pm0.43$ 
       & $87.35\pm0.96$ & $82.04\pm1.07$ & $90.81\pm0.82$ 
       & $\underline{85.47\pm0.64}$\\
HL-GNN & $92.71\pm0.98$ & $\underline{84.71\pm0.26}$ 
       & $\underline{93.24\pm0.45}$ & $\underline{86.53\pm0.67}$ 
       & $90.31\pm0.13$ & $85.59\pm0.08$ & $86.51\pm0.49$ & $79.88\pm1.35$ 
       & $\underline{91.14\pm0.56}$ & $84.31\pm1.79$ 
       & $\underline{92.25\pm0.77}$ & $\underline{85.42\pm0.62}$ 
       & $89.96\pm0.09$ & $84.73\pm1.02$\\
\midrule
AdaGCN  & $85.13\pm0.33$ & $81.59\pm0.15$ & $89.24\pm0.47$ & $83.20\pm0.80$          & $87.52\pm0.17$ & $82.62\pm0.93$ & $85.46\pm0.69$ & $81.12\pm1.04$          & $88.34\pm0.27$ & $81.42\pm0.24$ & $87.41\pm0.53$ & $81.04\pm0.83$          & $88.84\pm0.88$ & $82.56\pm1.55$\\
LGCL    & $89.34\pm0.41$ & $82.28\pm1.32$ & $88.15\pm1.63$ & $83.36\pm0.61$          & $85.31\pm0.06$ & $79.99\pm2.48$ & $84.79\pm1.32$ & $79.98\pm0.62$          & $86.41\pm0.51$ & $79.10\pm1.33$ & $85.93\pm0.68$ & $80.11\pm1.38$          & $87.25\pm2.01$ & $80.73\pm1.71$\\
LPFormer  & $\underline{92.82\pm0.29}$ & $84.05\pm1.30$ & $91.44\pm1.64$ 
          & $84.91\pm0.27$ & $\underline{92.07\pm1.72}$ 
          & $\underline{89.74\pm0.67}$ & $\underline{89.01\pm0.18}$ 
          & $82.51\pm1.08$ & $90.77\pm0.95$ & $\underline{85.46\pm1.32}$ 
          & $90.09\pm0.81$ & $84.41\pm0.25$ & $\underline{92.33\pm1.01}$ 
          & $85.10\pm0.66$\\
Graph2Feat  & $88.31\pm0.89$ & $82.07\pm1.37$ & $89.79\pm0.67$ 
            & $83.98\pm1.93$ & $90.22\pm0.97$ & $84.36\pm0.53$ 
            & $86.57\pm0.37$ & $80.73\pm0.18$ & $89.23\pm0.66$ 
            & $84.02\pm0.78$ & $88.62\pm1.69$ & $83.59\pm2.05$ 
            & $90.21\pm1.83$ & $84.71\pm0.61$\\
\midrule
GAATNet & $\mathbf{94.27\pm0.78}$ & $\mathbf{86.85\pm0.61}$ 
        & $\mathbf{95.28\pm0.08}$ & $\mathbf{88.61\pm0.18}$ 
        & $\mathbf{96.37\pm0.95}$ & $\mathbf{92.27\pm0.73}$ 
        & $\mathbf{92.69\pm1.06}$ & $\mathbf{87.53\pm0.89}$ 
        & $\mathbf{94.34\pm0.76}$ & $\mathbf{88.01\pm1.35}$
        & $\mathbf{95.05\pm0.64}$ & $\mathbf{89.86\pm0.81}$
        & $\mathbf{95.48\pm0.93}$ & $\mathbf{90.27\pm0.27}$\\
\bottomrule
\end{tabular}
}
\label{tab: results}
\end{table*}

Specifically, compared to the best-performing baseline, GAATNet achieves an overall improvement in both AUC and F1 scores. For example, on the small datasets Cora and Citeseer, GAATNet improved the AUC by 1.45\% and 2.04\% and the F1 score by 2.14\% and 2.08\%, respectively. On the medium datasets Facebook and ChCh-Miner, GAATNet showed more significant advantages, achieving improvements of 4.3\%, 3.68\% in AUC, and 2.53\%, 4.6\% in F1 score, respectively. On the more challenging large datasets, Coauthor-CS, Amazon Computers, and Email-Enron, GAATNet shows excellent performance with an AUC of 94.34\%, 95.05\%, and 95.48\%, and an F1 score of 88.01\%, 89.86\%, and 90.27\%, respectively. Compared to the best baseline algorithms, GAATNet improved the AUC by 3.2\%, 2.8\%, and 3.15\%, and the F1 score by 2.55\%, 4.44\%, and 4.8\%, respectively. These results show that GAATNet not only adapts well to small datasets but also exhibits remarkable robustness and generalizability when dealing with more complex and sparse large graph data.

HL-GNN shows the second-best performance on small datasets due to its reliance on explicit features and local structural information for feature modeling. In cases where the feature information of the graph is very sparse, HL-GNN is limited to local explicit features and cannot model the global sparse structure. LPFormer shows the second-best performance on medium datasets. Compared with HL-GNN, LPFormer improves results by incorporating an adaptive graph transformer. It can learn link-specific, multi-factor pairwise encodings via attention, allowing it to capture informative connectivity patterns that emerge as edge density increases. However, these methods often lead to inconsistent performance on large graph datasets. HL-GNN is limited by its reliance on explicit, local features and therefore struggles to capture global structural information. Despite LPFormer's use of adaptive pairwise encodings, it does not employ transfer learning and is thus more susceptible to encoding overfitting and noise on large graphs.

In contrast, GAATNet introduces the graph transformer and transfer learning, which combine attention mechanisms. These mechanisms are capable of capturing long-range information and effectively modeling the global features within the graph. This measure allows GAATNet to maintain strong performance on both sparse and large graphs. Additionally, GAATNet utilizes transfer learning to extract rich feature information from the source graph, which makes up for the lack of sparse features in the target graph.

\textbf{Comparison under imbalanced settings}. To evaluate link prediction performance under imbalanced settings, we conducted experiments on three small-to-medium datasets. We tested two negative: positive ratios, 10:1 and 20:1, and used AUC and Average Precision (AP) as evaluation metrics. All other experimental settings are the same as in the balanced experiments. Results are shown in Table \ref{tb: AUC_AP_Comparison}. It can be observed that GAATNet achieves the best performance at both imbalanced ratios, indicating it can robustly identify positive links when negatives dominate. On small datasets, the second-best result is not tied to a single method. Ranks are highly competitive, suggesting greater variability and sensitivity to randomness and sampling when positive examples are very scarce. By contrast, on medium datasets, LPFormer obtains the second-best performance in most cases. Both GAATNet and LPFormer use the transformer module, which suggests that attention mechanisms can still learn discriminative embeddings for positive samples under imbalanced settings. Specifically, GAATNet’s pretrain–finetune strategy and the use of distant neighbor information as an attention bias help amplify useful positive information and suppress noise. The self-adapter reduces overfitting during fine-tuning and improves stability with limited samples. Overall, GAATNet demonstrates strong representation-learning ability for imbalanced link prediction.

\begin{table*}[t]
\centering
\caption{Results (\%) for link prediction under imbalanced settings on three datasets. \textbf{Bold} numbers denote the best performances and \underline{underline} numbers denote the second-best performances.}
\resizebox{\textwidth}{!}{
\begin{tabular}{lccccccccccccc}
\toprule
& \multicolumn{2}{c}{Cora (10:1)} & \multicolumn{2}{c}{Cora (20:1)} & \multicolumn{2}{c}{Facebook (10:1)} & \multicolumn{2}{c}{Facebook (20:1)} & \multicolumn{2}{c}{ChCh-Miner (10:1)} & \multicolumn{2}{c}{ChCh-Miner (20:1)}\\
\cmidrule(lr){2-3} \cmidrule(lr){4-5} \cmidrule(lr){6-7} \cmidrule(lr){8-9} \cmidrule(lr){10-11} \cmidrule(lr){12-13}
& \multicolumn{1}{c}{AUC} & \multicolumn{1}{c}{AP} 
& \multicolumn{1}{c}{AUC} & \multicolumn{1}{c}{AP} 
& \multicolumn{1}{c}{AUC} & \multicolumn{1}{c}{AP} 
& \multicolumn{1}{c}{AUC} & \multicolumn{1}{c}{AP} 
& \multicolumn{1}{c}{AUC} & \multicolumn{1}{c}{AP} 
& \multicolumn{1}{c}{AUC} & \multicolumn{1}{c}{AP}\\
\midrule
GCN   & $56.82\pm1.36$ & $57.71\pm0.98$ & $56.44\pm0.79$ & $56.57\pm1.23$ 
      & $59.89\pm0.71$ & $56.14\pm2.36$ & $53.92\pm1.95$ & $55.16\pm0.68$ 
      & $59.38\pm1.38$ & $60.25\pm0.81$ & $57.75\pm0.93$ & $60.63\pm0.39$\\
GAT   & $63.36\pm0.57$ & $66.97\pm0.88$ & $65.11\pm1.26$ & $65.85\pm2.07$ 
      & $71.67\pm1.87$ & $70.29\pm1.56$ & $73.48\pm2.14$ & $73.76\pm0.56$ 
      & $68.31\pm0.80$ & $71.55\pm1.38$ & $70.42\pm0.66$ & $70.93\pm0.19$\\
\midrule                    
SEAL  & $76.16\pm1.13$ & $79.25\pm0.85$ & $78.99\pm0.39$ & $80.46\pm0.48$ 
      & $79.73\pm1.37$ & $79.84\pm1.08$ & $83.29\pm0.77$ & $82.48\pm1.36$ 
      & $81.38\pm0.89$ & $83.57\pm1.57$ & $79.92\pm0.78$ & $77.67\pm0.43$\\
SIEG  & $84.45\pm1.34$ & $84.76\pm2.06$ & $79.89\pm2.15$ & $78.31\pm0.90$ 
      & $83.72\pm0.17$ & $80.42\pm0.48$ & $79.75\pm1.89$ & $77.29\pm0.58$ 
      & $80.67\pm1.37$ & $82.31\pm0.44$ & $81.55\pm0.11$ & $81.93\pm1.55$\\
HL-GNN  & $84.17\pm0.85$ & $83.85\pm1.37$ & $\underline{84.42\pm1.22}$ 
        & $\underline{84.76\pm0.88}$ & $87.93\pm0.41$ & $87.44\pm0.81$ 
        & $83.51\pm0.94$ & $\underline{84.36\pm0.05}$ & $86.41\pm0.57$ 
        & $84.71\pm1.09$ & $85.14\pm1.25$ & $\underline{86.36\pm0.85}$\\
\midrule

AdaGCN  & $84.91\pm0.77$ & $\underline{86.54\pm1.27}$ & $81.56\pm0.87$ 
        & $82.49\pm0.83$ & $86.98\pm0.78$ & $87.35\pm1.81$ & $82.94\pm0.56$ & $83.65\pm0.95$ & $83.43\pm1.67$ & $84.96\pm0.15$ & $82.15\pm1.23$ & $82.12\pm1.88$\\
LGCL    & $83.88\pm0.67$ & $84.32\pm1.44$ & $79.74\pm1.92$ & $81.36\pm0.26$          & $84.61\pm1.72$ & $83.23\pm0.55$ & $\underline{84.78\pm0.89}$ 
        & $80.83\pm1.75$ & $87.52\pm1.16$ & $85.47\pm0.67$ & $83.81\pm0.92$ & $83.68\pm0.36$\\
LPFormer & $\mathbf{88.76\pm0.45}$ & $85.37\pm1.42$ & $83.90\pm0.99$ 
         & $84.24\pm1.12$ & $\underline{88.07\pm1.46}$ 
         & $\underline{88.34\pm0.66}$ & $84.62\pm0.74$ & $83.70\pm0.91$ 
         & $\underline{89.69\pm1.16}$ & $\underline{88.46\pm0.36}$ 
         & $\underline{87.17\pm0.51}$ & $85.79\pm0.47$\\
Graph2Feat  & $84.64\pm1.06$ & $83.72\pm0.21$ & $79.51\pm0.32$ 
            & $81.58\pm1.41$ & $85.45\pm1.49$ & $84.87\pm2.05$ 
            & $82.77\pm0.92$ & $82.18\pm1.97$ & $86.33\pm0.47$ 
            & $85.07\pm1.86$ & $86.95\pm0.57$ & $84.86\pm1.94$\\
\midrule
GAATNet & $\underline{88.23\pm0.21}$ & $\mathbf{89.17\pm1.79}$ 
        & $\mathbf{87.51\pm0.97}$ & $\mathbf{85.42\pm2.03}$ 
        & $\mathbf{90.03\pm1.97}$ & $\mathbf{91.44\pm0.07}$ 
        & $\mathbf{86.96\pm1.79}$ & $\mathbf{86.18\pm0.59}$ 
        & $\mathbf{91.83\pm1.46}$ & $\mathbf{91.13\pm0.66}$ 
        & $\mathbf{90.07\pm1.28}$ & $\mathbf{90.88\pm2.17}$\\
\bottomrule
\end{tabular}
}
\label{tb: AUC_AP_Comparison}
\end{table*}

\subsection{Ablation study}
In this section, our goal is to assess the effectiveness of each component in our method.

\textbf{Effect of pre-training network}. To evaluate the effectiveness of the pre-training component in link prediction, we designed three experimental strategies for comparison: GAATNet, $\text{GAATNet}_s$, and $\text{GAATNet}_t$. $\text{GAATNet}_s$ represents the model trained from scratch, where all network parameters are randomly initialized. $\text{GAATNet}_t$ refers to the model that uses Transformer-based pretrained parameters and fine-tuning based on it. GAATNet refers to the complete method we propose. It includes both pre-training and fine-tuning in a two-stage learning process. All experiments used the Amazon Computers dataset as the source graph for pretraining to ensure the consistency of the experiments.

\begin{figure*}[ht]
    \centering
    \includegraphics[clip,scale=0.22]{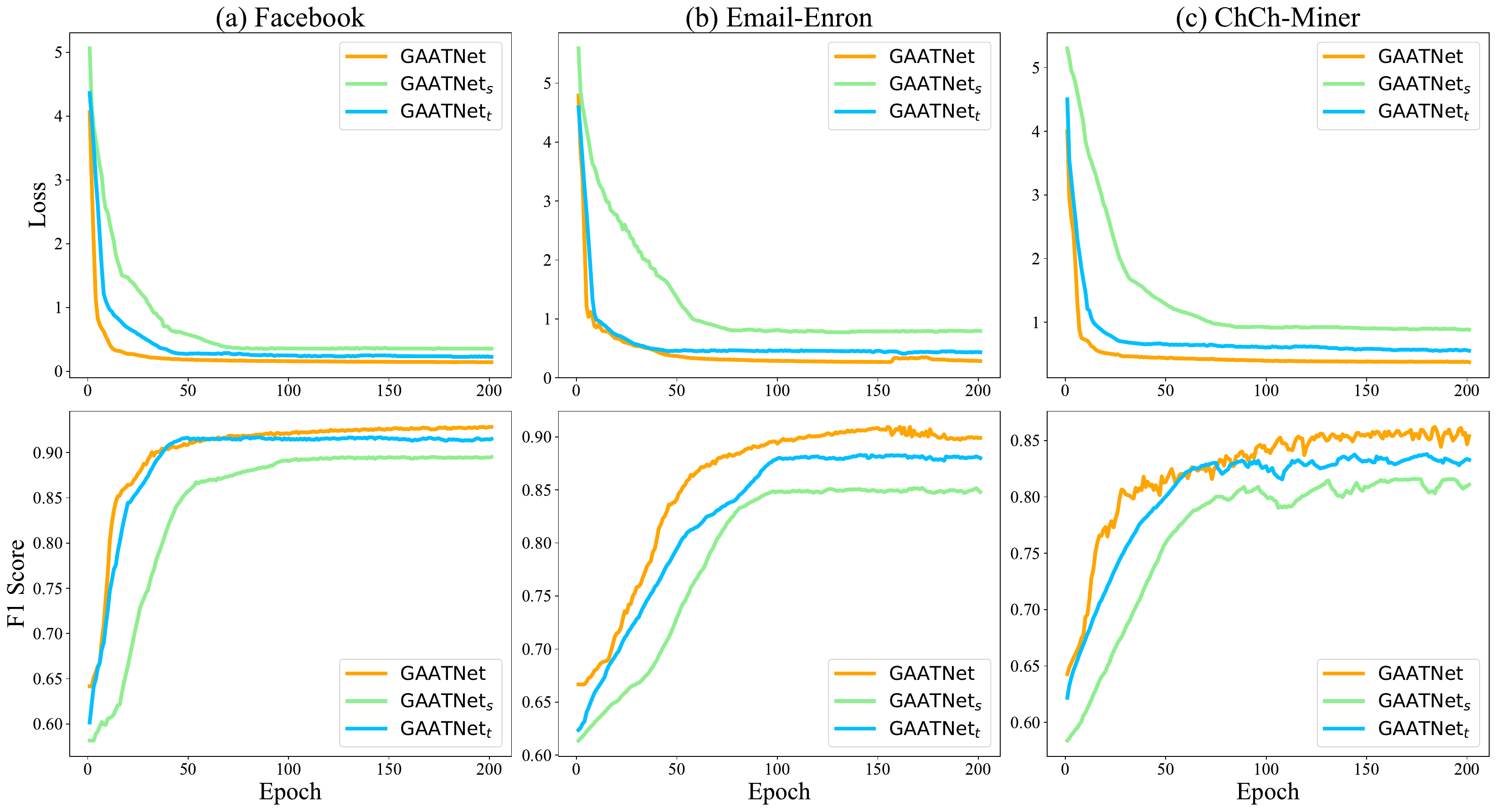}
    \caption{Effectiveness of the pre-training. $\text{GAATNet}_s$ represents a model trained from scratch without loading pretrained model parameters. $\text{GAATNet}_t$ refers to a model that loads Transformer-based pretrained parameters and fine-tunes based on it.}
    \label{fig: as_pre-training}
\end{figure*}

In the experiments, we chose the Facebook, Email-Enron, and ChCh-Miner datasets for evaluation and compared the loss and F1 score of the three strategies. The experimental results are shown in Fig. \ref{fig: as_pre-training}. By analyzing these results, we can draw several important conclusions. Firstly, the training process of $\text{GAATNet}_t$ and GAATNet significantly outperforms $\text{GAATNet}_s$. Especially in the early stages of training, the model with pretrained parameters quickly reaches higher performance, and the loss curve shows a more stable downward trend. However, $\text{GAATNet}_t$ performs slightly worse than GAATNet. This is due to the GAT and information from distant neighbors. These factors lead to node representations that contain more comprehensive information and more critical features. This indicates that pre-training provides a good initial state for the model, thus accelerating the learning process.

In summary, our experimental results validate the effectiveness of pre-training in link prediction, especially when dealing with large-scale graph data. GAATNet reduces loss faster during fine-tuning and shows greater improvement in F1 score. This indicates that the knowledge transferred from the source graph enhances model performance, even with sparse target graph data. Pre-training enables the model to converge to the optimal solution more quickly. This is crucial for reducing computational resource consumption and speeding up model deployment.

\begin{figure}[ht]
    \centering
    \includegraphics[clip,scale=0.2]{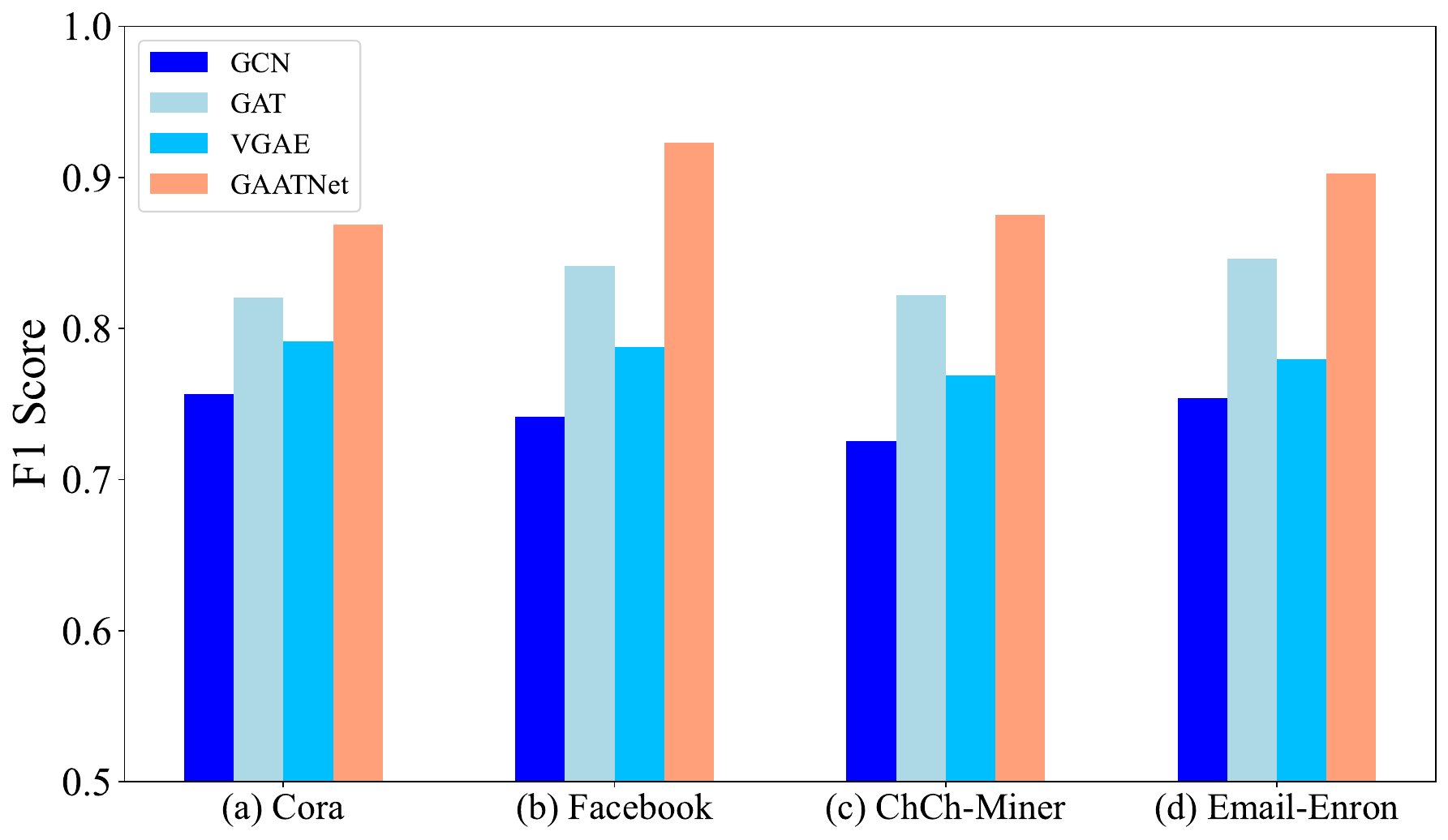}
    \caption{Effectiveness of the fine-tuning.}
    \label{fig: as_fine-tuning}
\end{figure}

\textbf{Effect of fine-tuning network}. To evaluate the effectiveness of the fine-tuning component, we replaced the fine-tuning network with GCN, GAT, and VGAE. In the experiments, we first pretrain on the Amazon Computers dataset. Then, we use the pre-trained parameters to fine-tune GCN, GAT, and VGAE. Finally, we compare their performance with our method. The fine-tuning is performed on four datasets of different scales, Cora, Facebook, ChCh-Miner, and Email-Enron, which are evaluated using the F1 score. The experimental results are shown in Fig. \ref{fig: as_fine-tuning}. It can be observed that our method is optimal, which proves the effectiveness of our fine-tuning network.

From Fig. \ref{fig: as_fine-tuning} (a), we can see that on the small dataset Cora, the performance gap between the four methods is not significant, and the improvement of GAATNet is relatively limited. However, from Fig. \ref{fig: as_fine-tuning} (b), Fig. \ref{fig: as_fine-tuning} (c), and Fig. \ref{fig: as_fine-tuning} (d), the differences between the methods increase. This is because the other three methods are not able to effectively capture node representations in sparse graphs on large datasets. GAATNet shows a more significant advantage on these datasets.

Further analysis reveals that GAT also uses an attention mechanism and achieves the second-best performance. However, there is still a gap compared to GAATNet. This is because, in addition to using the attention mechanism, our method introduces improvements at multiple levels, making node embeddings more accurate and efficient. In contrast, GCN and VGAE perform relatively poorly. The GCN relies only on simple graph convolution operations, which cannot effectively capture the complex details of graph structure, especially on large sparse graphs. The generative model and reconstruction task design of VGAE limit its adaptability during fine-tuning, thus affecting its performance on link prediction. In conclusion, our approach is optimized in multiple aspects to significantly improve performance. The experimental results validate the effectiveness of our fine-tuning network, especially when dealing with large sparse graphs, where our method brings more substantial performance improvements.

\textbf{Effect of dual-stage transfer learning}. To evaluate the effectiveness of our two-stage transfer learning component versus traditional transfer learning, we compared $\text{GAATNet}$ with $\text{GAATNet}_{Trad}$ in the Facebook dataset. $\text{GAATNet}_{Trad}$ directly fine-tunes the pretrained model, updating only the output layer parameters. The experimental results show that $\text{GAATNet}$ significantly improves both accuracy and efficiency. Specifically, $\text{GAATNet}_{Trad}$ achieves an AUC of 83.93\% with an average time consumption of 19.11 seconds per epoch. In contrast, $\text{GAATNet}$ attains a substantially higher AUC of 96.37\% while reducing the per-epoch time to only 2.25 seconds. This demonstrates that GAATNet’s two-stage, parameter-efficient transfer learning strategy improves accuracy by reducing overfitting and using the lightweight self-adapter module. At the same time, it greatly reduces computational cost by freezing most pretrained parameters and updating only a small portion.

\begin{figure}[ht]
    \centering
    \includegraphics[clip,scale=0.16]{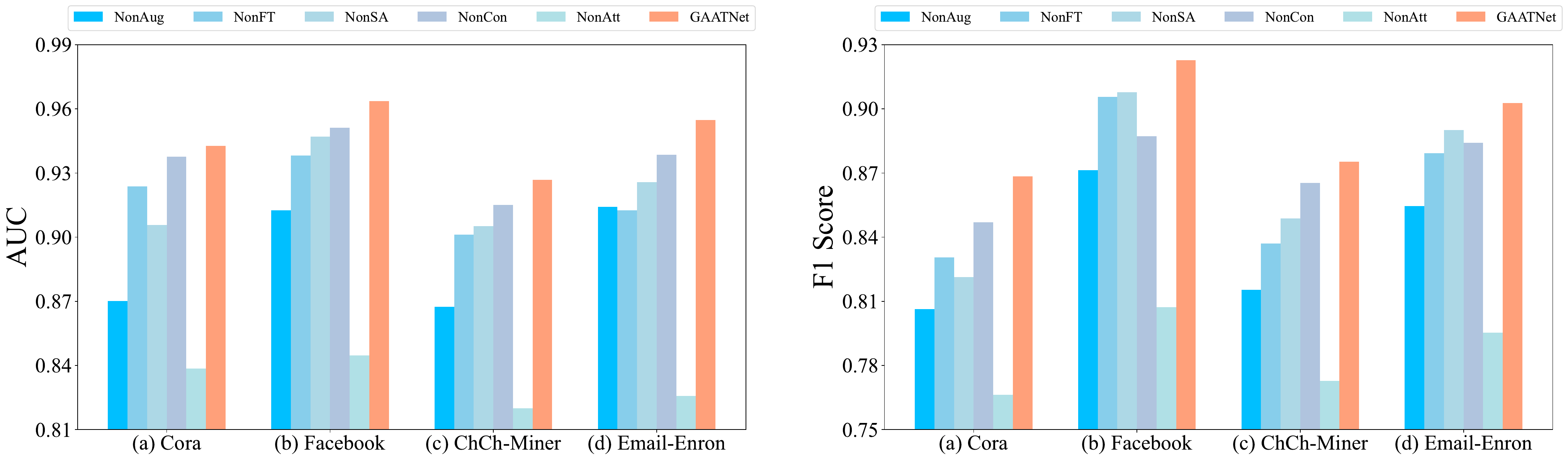}
    \caption{Effectiveness of different training strategies.}
    \label{fig: as_auc_f1_comparison}
\end{figure}

\textbf{Effect of key components}. We conducted an ablation study by designing different training strategies to assess the effectiveness of each key component in GAATNet. The strategies include NonAug, NonFT, NonSA, NonCon, NonAtt, and GAATNet. NonAug indicates the removal of the diffusion data augmentation strategy, which uses only the original data for training. NonFT represents the removal of the feature transformer module in the pretraining network. NonSA denotes the removal of the self-adapter module in the fine-tuning network. NonCon represents the removal of the contrastive loss, as shown in Equation (\ref{eq12}). NonAtt denotes replacing the attention-based modules in GAATNet with mean pooling. Specifically, the attention layers in the feature transformer encoder and in the GAT are replaced by mean pooling. We performed experiments on four datasets of different scales, and the results are shown in Fig. \ref{fig: as_auc_f1_comparison}.

The experimental results show that GAATNet achieves the highest AUC and F1 score and performs optimally. It can be observed that NonAug has the second-lowest performance. This model relies only on raw data for training and ignores diffusion data augmentation. As a result, the node representations contain less information and are sparser, which leads to the worst performance. On the small dataset Cora, NonFT performs better than NonSA. This is because, on smaller datasets, the Transformer model is relatively complex and tends to lead to overfitting when the number of edges is small. In contrast, on medium and large datasets, NonSA outperforms NonFT. This is because, during pretraining, the feature transformer module learns crucial information from complex sparse graphs. This helps the model better adapt to large graph learning during fine-tuning, thus improving performance. NonCon shows the second-best performance, as it retains the complete pretraining and fine-tuning network. This makes it more adaptable to datasets of different scales and learns more comprehensive node embeddings. Compared to GAATNet, the average performance of NonCon is only 1.145\% lower. By introducing contrastive loss, the distance between positive node pairs is reduced. This makes the embeddings of different node pairs more distinguishable, further improving GAATNet’s performance. NotAtt, which lacks attention layers, shows the worst performance. The attention mechanism adaptively reweights neighbors or features, suppressing source-specific or noisy signals and thus enabling the model to learn more transferable and informative representations. This experimental result shows that all components of GAATNet are effective and contribute to performance improvements at different levels.

\textbf{Effect of similarity score calculation}. To validate the effectiveness of similarity score computation, we compared $\text{GAATNet}$ with $\text{GAATNet}_{dot}$, where $\text{GAATNet}_{dot}$ computes similarity by vector dot product. Results are reported in Table \ref{tb: score_calc}. The Euclidean-distance-based, nonlinear similarity measure used in GAATNet achieves the highest AUC across all datasets. These findings indicate that nonlinear similarity measures are generally more effective than linear ones.

\begin{table}[t]
\scriptsize
\setlength{\tabcolsep}{3pt} 
\centering
\caption{The AUC values (\%) of similarity score computation across all datasets. $\text{GAATNet}_{dot}$ denotes results computed using the vector dot product. The highest value is marked in \textbf{bold}.}
\label{tb: score_calc}
\begin{tabular}{lccccccc}
\toprule
 & Cora & Citeseer & Facebook & ChCh & Co-CS & Computers & Email\\
\midrule
$\text{GAATNet}_{dot}$ & 91.23 & 90.44 & 93.36 & 89.74 & 93.28 & 92.74 & 93.79 \\
$\text{GAATNet}$ & \textbf{94.27} & \textbf{95.28} & \textbf{96.37} & \textbf{92.69} & \textbf{94.34} & \textbf{95.05} & \textbf{95.48} \\
\bottomrule
\end{tabular}
\end{table}

\subsection{Parameter sensitivity analysis} 

\begin{figure*}[ht]
    \centering
    \includegraphics[clip,scale=0.2]{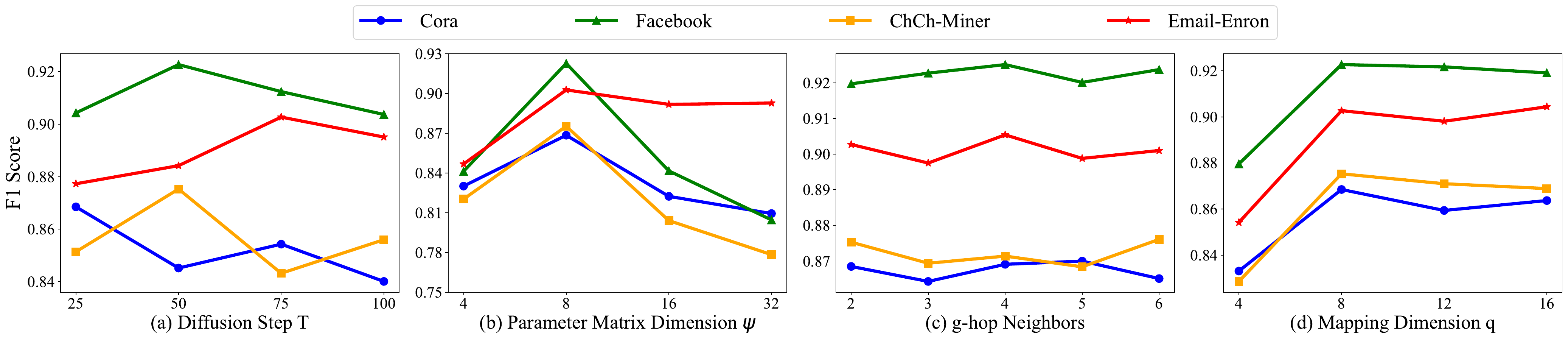}
    \caption{Parameter sensitivity studies of four key hyperparameters. F1 score comparison of GAATNet's performance on four datasets.}
    \label{fig: psa_FourColsFourPara}
\end{figure*}

To analyze the sensitivity of GAATNet's performance to hyperparameters, we conducted experiments on four key hyperparameters: the diffusion steps $T$, the dimension of the learnable parameter matrix $\psi$ in Eq. \ref{eq10}, the $g$-hop distant neighbors, and the mapping dimension $q$ of the self-adapter module. The experimental results are shown in Fig. \ref{fig: psa_FourColsFourPara}.

From Fig. \ref{fig: psa_FourColsFourPara} (a), we can see that in the experiments on diffusion steps, the small dataset achieved optimal performance at 25 steps, the medium datasets performed best at 50 steps, and the large dataset required 75 steps to reach the best results. The number of diffusion steps determines the range of information propagation in the diffusion data augmentation process. Smaller diffusion steps are more suitable for small graphs, while larger graphs require more steps so that nodes can access more distant neighbors and gather more comprehensive information. However, increasing the diffusion steps also significantly increases computational time. Therefore, to balance performance and computational resources, we chose a compromise of 50 diffusion steps for data augmentation.

Fig. \ref{fig: psa_FourColsFourPara} (b) illustrates the results for the dimension of the learnable parameter matrix $\psi$. When the dimension is set to 8, the performance is optimal for all datasets. This hyperparameter is associated with the dimension of node embeddings output by the attention enhancement output layer. Its size directly influences the amount of key information contained in the embeddings. It can be observed that this parameter has the most significant impact on model performance and is either too large or too small, potentially degrading performance. When the dimension is set to 4, it is too small, leading to insufficient feature information. This makes it difficult for the model to effectively determine link generation, thus leading to performance degradation. In contrast, when the dimension is set to 16 or 32, it may introduce redundant or irrelevant information, making it harder to focus on critical information, which also reduces performance. At a dimension of 8, it is just enough to cover the key information and effective link prediction. Furthermore, it can also be observed that on the large dataset Email-Enron, the performance does not decline significantly despite dimension settings of 16 or 32. This suggests that on large datasets, moderately increasing the dimension may not negatively impact performance. Therefore, we selected a dimension of 8 as the optimal setting for our method.

Fig. \ref{fig: psa_FourColsFourPara} (c) shows the results for $g$-hop distant neighbors. This hyperparameter has a relatively small impact, with performance differences in each dataset under different $g$ settings being less than 1\%. This is because, in graph structure data, most nodes are no longer likely to encounter new neighbors after 2 or 3 hops. When $g$ is set to 4, 5, or 6, most of the extracted neighbor information still comes from nodes within the 2- or 3-hop range, with only partial additional information contributed by nodes in the 4- to 6-hop range. Increasing the $g$-hop count provides limited improvement to performance while introducing excessive redundant computation and increased time complexity when extracting subgraphs. Therefore, in our method, $g$ is set to 2 or 3, which effectively aggregates neighbor information while reducing computational costs.

Fig. \ref{fig: psa_FourColsFourPara} (d) illustrates the results for the mapping dimension of the self-adapter module. When the dimension increases from 4 to 8, performance improves across all datasets. However, when the dimension is further increased, performance remains unchanged or improves slightly. This is similar to the dimension of the learnable parameter matrix; when the dimension is too small, critical information cannot be fully captured, which leads to degraded performance. However, since the self-adapter module performs feature enhancement through projection and fusion, the information loss is minimal. Therefore, increasing the dimension excessively does not significantly change performance but instead introduces additional learnable parameters, which reduces computational efficiency. Therefore, in our method, the mapping dimension of the self-adapter module is set to 8, ensuring both high performance and computational efficiency.

\begin{figure}[ht]
    \centering
    \includegraphics[clip,scale=0.24]{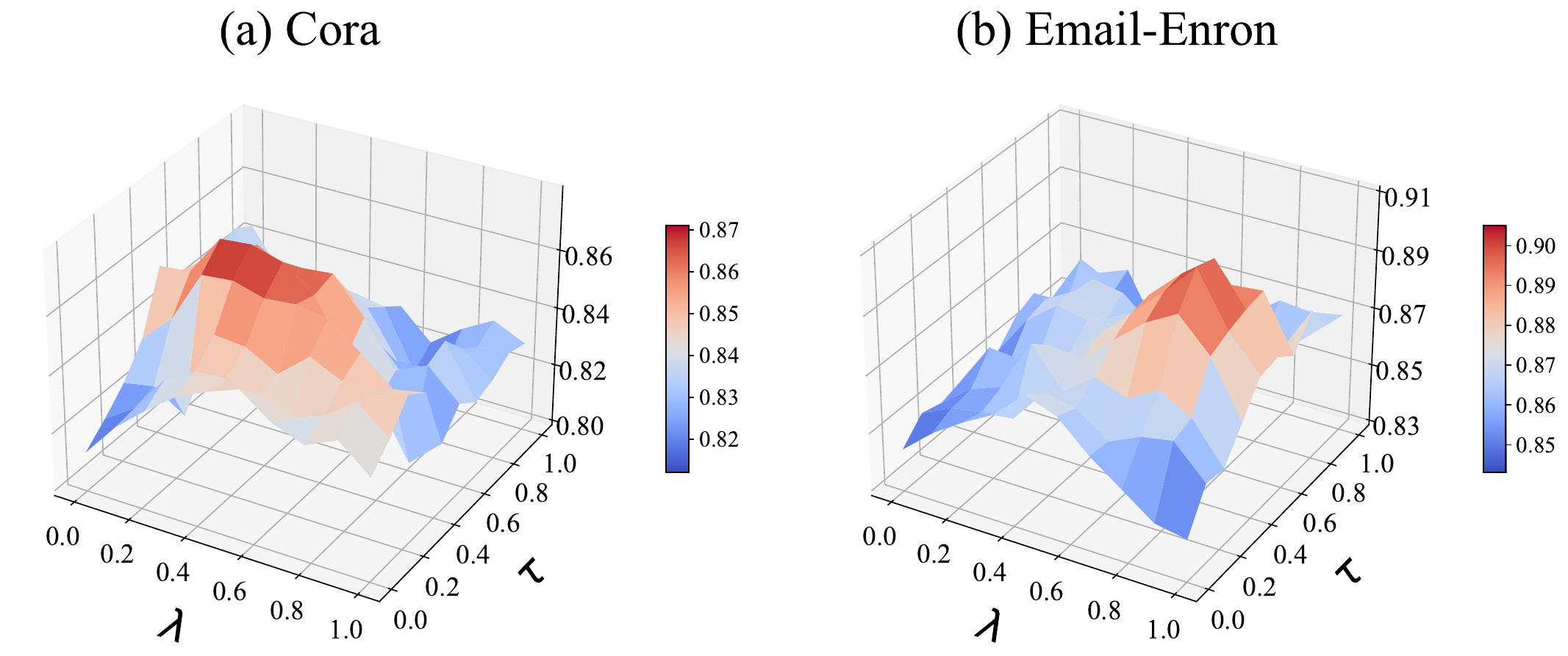}
    \caption{Parameter sensitivity studies of weight parameter $\lambda$ and temperature parameter $\tau$ on Cora and Email-Enron datasets.}
    \label{fig: psa_ConLoss3DPlot}
\end{figure}

To evaluate the impact of contrastive loss on model performance, we conducted an experimental analysis of the hyperparameters $\lambda$ in Eq. \ref{eq14} and $\tau$ in Eq. \ref{eq12} on the Cora and Email-Enron datasets. The results are shown in Fig. \ref{fig: psa_ConLoss3DPlot}. The results in Fig. \ref{fig: psa_ConLoss3DPlot} (a) show that the model achieves optimal performance on the Cora dataset when $\lambda$ is set to 0.4 or 0.5 and $\tau$ is set to 0.5. On small datasets like Cora, when $\lambda$ is small, the impact of the contrastive loss is minimal, and the model relies more on the link prediction loss. This leads the model to focus on the label information of the data. However, due to the limited number of samples, the model may learn noisy data, which results in overfitting. When $\lambda$ is large, the model emphasizes contrastive loss between positive and negative node pairs, while reducing its reliance on label information. However, for small graphs, which are denser compared to large graphs, excessively weakening the influence of label information may prevent the model from fitting the task well. This leads to performance degradation. On the other hand, $\tau$ controls the measure of similarity between samples. When $\tau$ is too small, the model becomes better at distinguishing between similar and dissimilar samples. However, this also makes the model too sensitive to noise in the data, which may lead to overfitting. In contrast, when $\tau$ is too large, the model may not be able to effectively distinguish between samples, leading to poor generalization. Therefore, on small datasets, the optimal settings for $\lambda$ and $\tau$ are moderate, with $\lambda$ set to 0.4 or 0.5 and $\tau$ set to 0.5.

From the results in Fig. \ref{fig: psa_ConLoss3DPlot} (b), we can see that the model achieves optimal performance on the Email-Enron dataset when $\lambda$ is set to 0.7 or 0.8 and $\tau$ is set to 0.5 or 0.6. This result is consistent with the previous analysis. When $\lambda$ is larger, the model focuses more on contrastive loss while reducing its reliance on label information. In the large sparse graphs, where label information is relatively limited, too much focus on label information can lead to underfitting. Therefore, more attention to contrastive loss between positive and negative node pairs is necessary. Additionally, the setting of $\tau$ still needs to be moderate, because a large $\lambda$ already makes the model heavily dependent on contrastive loss, and an overly large $\tau$ may trigger the risk of overfitting.

\subsection{Transferability analysis}

\begin{figure}[ht]
    \centering
    \includegraphics[clip,scale=0.2]{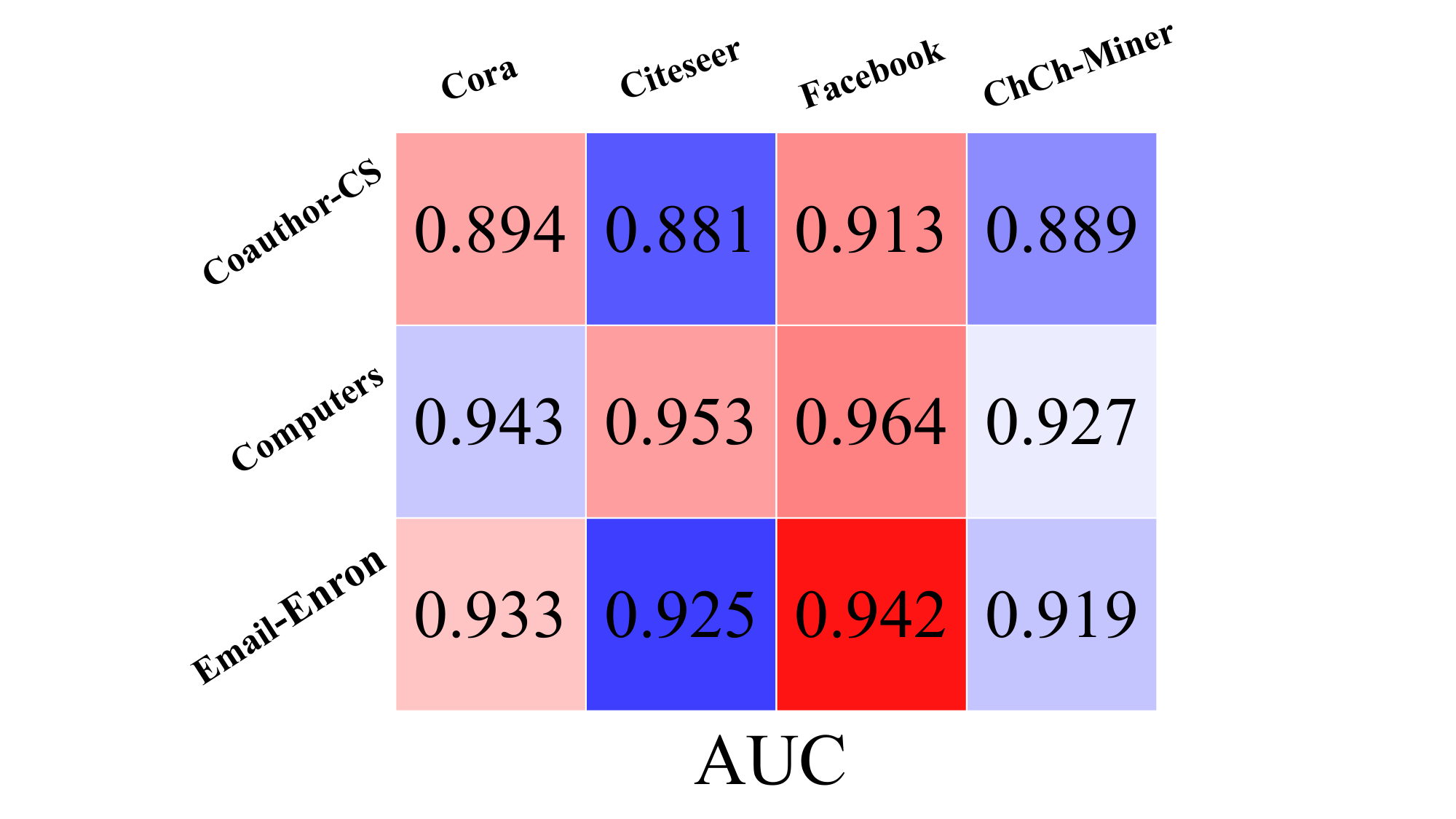}
    \caption{The AUC results of the transferability analysis.}
    \label{fig: ta_transfer}
\end{figure}

\textbf{Conditional analysis of transferability}. In this section, we analyze the importance of source graph density for knowledge transfer to smaller target graphs. We first pretrain models on three large datasets: Coauthor-CS, Computers, and Email-Enron. The pretrained models are then transferred to four small and medium datasets, and we evaluate their AUC performance. The experimental results are shown in Fig. \ref{fig: ta_transfer}. All large datasets can transfer knowledge effectively to the Facebook dataset, achieving the best performance, with Computers demonstrating the strongest transfer ability. For the small datasets, Cora and Citeseer obtain similar AUC values, whereas ChCh-Miner performs the worst. In combination with the statistics in Table \ref{table: datasets}, it can be observed that Computers has the highest graph density among the large datasets. A greater number of edges enables the learned node embeddings to capture richer information, thus providing more transferable knowledge. Moreover, Facebook is a social network graph, which is structurally closer to the large datasets, resulting in a higher AUC compared to the other three datasets. These findings indicate that higher graph density generally facilitates better knowledge transfer, while structural similarity between source and target graphs further enhances transfer performance.

\textbf{Transferability analysis across domains}. To evaluate GAATNet’s link prediction performance across different domains, we use two cross-domain datasets: Last.FM Music\footnote{\label{footmusic}http://www.lastfm.com} and Amazon Reviews (Baby)\textsuperscript{\ref{foot}}. Last.FM Music is a sparse heterogeneous graph with three node types and three edge types. The Baby dataset is a sparse interaction graph of Amazon reviews for baby products, where each review contains five rating levels. Dataset statistics are reported in Table \ref{tab: transfer_dataset}. For comparison on heterogeneous graphs, we use HGNN \cite{zhang2019heterogeneous}, HAN \cite{wang2019heterogeneous}, and LHGNN \cite{nguyen2023link} as baselines. To ensure fairness, we align their attention heads, hidden dimensions, and input/output sizes with those of GAATNet. The task is to predict edge types given a head and tail node, so we formulate it as a multi-class problem. Training utilizes cross-entropy loss, and evaluation employs Macro-F1 to reflect per-class performance. To study imbalance effects, we run two settings with negative: positive ratios of 1:1 and 10:1. The “negative” denotes the non-existent edges and is treated as an extra class during training. Results are shown in Table \ref{tb: F1_across_domains}. It can be observed that GAATNet achieves the best performance on both cross-domain datasets. Because our model is primarily designed for binary link prediction, its performance on multi-class heterogeneous graph tasks is not as strong as in the binary setting. However, the results show that the model can still effectively learn node embeddings in heterogeneous graphs, demonstrating good representational power and generalization ability.

\begin{table}[ht]
\small
\centering
\caption{Statistical overview of the datasets.}
\label{tab: transfer_dataset}
\begin{tabular}{lccccccc}
\toprule
 & \#Nodes & \#Edges & \#Features & \#Node types & \#Edge types & \#Avg (degree) \\
\midrule
Last.FM & 31,470 & 141,841 & 256 & 3 & 3 & 2.70      \\
Baby    & 130,777 & 176,186 & 256 & 2 & 5 & 9.01     \\
\bottomrule
\end{tabular}
\end{table}


\begin{table}[t]
\small
\centering
\caption{Macro-F1 (\%) results for link prediction across different domains. \textbf{Bold} numbers denote the best performances.}
\label{tb: F1_across_domains}
\begin{tabular}{lcccc}
\toprule
& Last.FM (1:1) & Last.FM (10:1) & Baby (1:1) & Baby (10:1)\\
\midrule
HGNN   & 67.44 & 65.40 & 65.36 & 62.94\\
HAN    & 73.51 & 72.62 & 74.15 & 70.46\\
LHGNN  & 76.18 & 74.82 & 71.49 & 69.83\\
\midrule                    
GAATNet & \textbf{81.68} & \textbf{77.26} & \textbf{79.17} & \textbf{76.45}\\
\bottomrule
\end{tabular}
\end{table}

\subsection{Efficiency analysis}

\textbf{Time efficiency}. To evaluate the time efficiency of GAATNet, we show the training time consumption per epoch for each method in Table \ref{tb: time_eff}, where $\text{GAATNet}_{p}$ and $\text{GAATNet}_{f}$ represent the training time at the pertaining and fine-tuning stages, respectively. The results show that GAATNet has relatively low time consumption during the fine-tuning stage, which is reduced by half on average compared to the pre-training stage. Although the traditional GNN models achieve the fastest training speed, their limited learning capacity results in lower accuracy. In contrast, SEAL requires converting the graph into a line graph for training, which significantly increases computational costs and results in the slowest training speed. LGCL is an improved variant of SEAL and thus incurs a similar data-processing cost. Additionally, the experiments have shown that the training time of a model is closely related to the number of nodes. The higher the number of nodes, the longer it takes for the model to update the node embeddings.

\begin{table}[t]
\scriptsize
\setlength{\tabcolsep}{3pt}
\centering
\caption{The average time consumption per epoch for different approaches (in seconds). $\text{GAATNet}_{p}$ and $\text{GAATNet}_{f}$ represent the pretraining and fine-tuning stages. The shortest and second shortest times are marked in \textbf{bold} and \underline{underline}, respectively.}
\label{tb: time_eff}
\begin{tabular}{lccccccc}
\toprule
 & Cora & Citeseer & Facebook & ChCh & Co-CS & Computers & Email\\
\midrule
GCN & \textbf{0.05} & \textbf{0.07} & \underline{0.85} & \textbf{0.43} & \textbf{8.67} & \textbf{4.32} & \textbf{6.94} \\
GAT & \underline{0.08} & \underline{0.11} & 1.03 & \underline{0.51} & \underline{9.55} & \underline{5.81} & \underline{7.72} \\
SEAL & 29.3 & 27.8 & 450.3 & 240.8 & 4980 & 2271 & 3108 \\
SIEG & 2.74 & 3.02 & 19.6 & 16.4 & 154.5 & 120.1 & 143.2 \\
HLGNN & 0.13 & 0.18 & 0.97 & 0.59 & 10.48 & 6.38 & 8.42 \\

AdaGCN & 4.33 & 4.96 & 32.04 & 25.83 & 410.6 & 213.2 & 304.7 \\
LGCL & 27.82 & 26.64 & 397.5 & 231.3 & 3154 & 1842 & 2366 \\
LPFormer & 3.64 & 4.18 & 28.63 & 18.92 & 267.4 & 174.9 & 228.8 \\
Graph2Feat & 1.35 & 1.61 & 10.45 & 6.79 & 81.06 & 55.37 & 68.44 \\
\midrule
$\text{GAATNet}_{p}$ & 0.48 & 0.74 & 1.36 & 1.25 & 31.03 & 16.32 & 20.56 \\
$\text{GAATNet}_{f}$ & 0.21 & 0.32 & \textbf{0.63} & 0.57 & 13.45 & 7.29 & 9.29 \\
\bottomrule
\end{tabular}
\end{table}

\textbf{Parameter and memory efficiency}. The number of parameters and memory consumption are critical metrics in the model deployment process because they directly affect deployment costs. Therefore, we compared GAATNet with other methods in terms of the number of trainable parameters and the total memory consumption on the Cora dataset. The results are shown in Table \ref{tb: parameters_memory}. It can be observed that GAATNet achieves the best parameter efficiency while consuming the least memory. Among the methods, SEAL requires more parameters and memory due to the need to train on line graphs, which involve a larger number of edges. Furthermore, it is worth noting that the number of parameters and memory consumption of GAATNet during the fine-tuning stage are significantly lower than those during the pretraining stage. This demonstrates that the self-adapter module plays a crucial role, not only in effectively reducing computational costs but also in providing outstanding predictive performance, which exhibits good scalability and time efficiency.

\begin{table}[t]
    \footnotesize
    \setlength{\tabcolsep}{2.4pt}
    \centering
    \caption{Comparison of the number of trainable parameters (\#Parameters) and the total memory consumption on the Cora dataset (\#Memory). The smallest values are marked in \textbf{bold}, and the second smallest is \underline{underline}.}
    \label{tb: parameters_memory}
\begin{tabular}{lcccccccccccc}
\toprule
& GCN & GAT & SEAL & SIEG & HL-GNN & AdaGCN & LGCL & LPFormer & Graph2Feat & $\text{GAATNet}_{p}$ & $\text{GAATNet}_{f}$ \\
\midrule
\#Parameters (k)
& 312 & 384 & 2300 & 915 & 374 & 1220 & 1975 & 986 & 781 & \underline{208} & \textbf{44} \\

\#Memory (MB)
& 410 & 443 & 1062 & 953 & 661 & 933 & 1293 & 962 & 856 & \underline{271} & \textbf{57} \\
\bottomrule
\end{tabular}
\end{table}

\section{Conclusion} \label{sec: conclusion}

In this study, we propose a novel link prediction method, the Graph Attention Adaptive Transfer Network (GAATNet). GAATNet effectively applies a transfer learning strategy by pretraining on large-scale graph data and fine-tuning on small, sparse graphs. GAATNet first extracts node embedding matrices using node2vec. During the pretraining stage, it employs diffusion-based data augmentation, a graph multi-head attention module, and a feature transformer module. The outputs are enhanced with attention mechanisms, and similarity scores are computed to represent prediction probabilities. A lightweight self-adapter module is introduced to enable efficient computation in the fine-tuning stage. Experimental results on seven public datasets demonstrate that GAATNet outperforms existing methods in both predictive performance and efficiency. This highlights the significant potential of transfer learning for link prediction tasks.

Additionally, there are several directions that we should continue to explore in future work. For example, extending the framework to dynamic, spatio-temporal, or heterogeneous graphs could better adapt to the demands of graph topology applications. Further research could focus on developing more lightweight pretraining networks suitable for deployment in scenarios with limited computational and memory resources. What's more, adopting more efficient graph augmentation methods may help alleviate underfitting issues caused by sparse graph data. At the same time, consider incorporating cutting-edge technologies from other fields, such as large language models, knowledge distillation, and reinforcement learning, to further expand the application scope of link prediction techniques and provide more ideas and directions for future research.

\section*{Acknowledgment}

This research was supported in part by the National Natural Science Foundation of China (No. 62272196), Guangzhou Basic and Applied Basic Research Foundation (No. 2024A04J9971), and the Beijing Natural Science Foundation (No. 4254079).

\bibliographystyle{ACM-Reference-Format}
\bibliography{main.bib}

\end{document}